\documentclass{article}
\usepackage{iclr2023_conference,times}

\usepackage{hyperref}
\usepackage{url}

\usepackage{graphicx}
\usepackage{multirow}
\usepackage{makecell}
\usepackage{booktabs}
\usepackage{wrapfig}

\usepackage{amsmath}

\usepackage{amssymb}

\usepackage[linesnumbered,ruled,vlined]{algorithm2e}

\SetKwInput{kwInput}{Input}
\SetKwInput{kwOutput}{Output}
\SetKwComment{Comment}{//}{}

\usepackage{subcaption}
\usepackage[export]{adjustbox}

\newcommand{\red}[1]{\textcolor[RGB]{255, 49, 49}{#1}}

\definecolor{rebuttal}{RGB}{255, 155, 0}

\makeatletter
\def\blfootnote{\xdef\@thefnmark{}\@footnotetext}
\makeatother

\iclrfinalcopy

\title{Revisiting the Assumption of Latent Separability for Backdoor Defenses}

\footnotetext[1]{The first two authors contributed equally to this paper.}
\footnotetext{
 Correspondence to: Xiangyu Qi, Tinghao Xie, Yiming Li, and Prateek Mittal.
}

\author{
Xiangyu Qi$^{1}$\footnotemark[1]~, Tinghao Xie$^{1}$\footnotemark[1]~, Yiming Li$^{2}$, Saeed Mahloujifar$^{1}$, Prateek Mittal$^{1}$\\
 $^{1}$Princeton University\\
 $^{2}$Tsinghua Shenzhen International Graduate School, Tsinghua University\\
\texttt{\{xiangyuqi,thx,sfar,pmittal\}@princeton.edu}; \texttt{li-ym18@mails.tsinghua.edu.cn}
}

\begin{document}

\maketitle

\begin{abstract}

Recent studies revealed that deep learning is susceptible to backdoor poisoning attacks. An adversary can embed a hidden backdoor into a model to manipulate its predictions by only modifying a few training data, \textit{without controlling the training process}. Currently, a tangible signature has been widely observed across a diverse set of backdoor poisoning attacks --- {models trained on a poisoned dataset tend to learn separable latent representations for poison and clean samples}. This latent separation is so pervasive that a family of backdoor defenses directly take it as a default assumption~(dubbed \emph{latent separability assumption}), based on which to identify poison samples via cluster analysis in the latent space. An intriguing question consequently follows: \emph{is the latent separation unavoidable for backdoor poisoning attacks}? This question is central to understanding whether the assumption of latent separability provides a reliable foundation for defending against backdoor poisoning attacks. In this paper, we design \textit{adaptive backdoor poisoning attacks} to present counter-examples against this assumption. Our methods include two key components: (1) a set of trigger-planted samples correctly labeled to their semantic classes~(other than the target class) that can regularize backdoor learning; (2) asymmetric trigger planting strategies that help to boost attack success rate~(ASR) as well as to diversify latent representations of poison samples. Extensive experiments on benchmark datasets verify the effectiveness of our adaptive attacks in bypassing existing latent separation based defenses. Our codes are available at \url{https://github.com/Unispac/Circumventing-Backdoor-Defenses}.

\end{abstract}

%\blfootnote{\textsuperscript{*} The first two authors contributed equally to this work.}

\section{Introduction}

%\vspace{-2mm}
Overparameterized deep neural network~(DNN) models can fit complex datasets perfectly and generalize well on $i.i.d.$ data distributions. However, the strong capacity of these models also render them susceptible to %adversarial attacks on their training data~\citep{jagielski2018manipulating,gu2017badnets,shafahi2018poison,zhao2017efficient} --- by excessively fitting those maliciously added/manipulated data samples, the resulting models can suffer from significant performance degradation or targeted mispredictions. 
%As one of the most typical examples, 
\textit{backdoor poisoning attacks}~\citep{gu2017badnets, Chen2017TargetedBA,turner2019label,li2022backdoor}. In a backdoor poisoning attack, an adversary only manipulates a small portion of the victim's training dataset. The victims will train their own model on the manipulated dataset and consequently get a backdoored model. Typically, the adversary will \textit{poison} the victim's dataset by injecting a small amount of {backdoor poison samples}, each of which contains a {backdoor trigger}~(e.g. a specific pixel patch) and is labeled to a specific {target class}. A DNN model trained on this poisoned dataset will be backdoored in that they tend to learn an artificial correlation between the backdoor trigger and the target class. These attacks are stealthy since backdoored models behave normally on natural samples and therefore users can hardly identify them.

%These attacks are stealthy and threatening, because backdoored models behave normally on natural samples but can misclassify a considerable amount of samples to the target class when the backdoor trigger is applied to these samples in test time.

\begin{figure}
\centering
\vspace{-2em}
\begin{subfigure}{0.18\textwidth}
    \includegraphics[width=\textwidth]{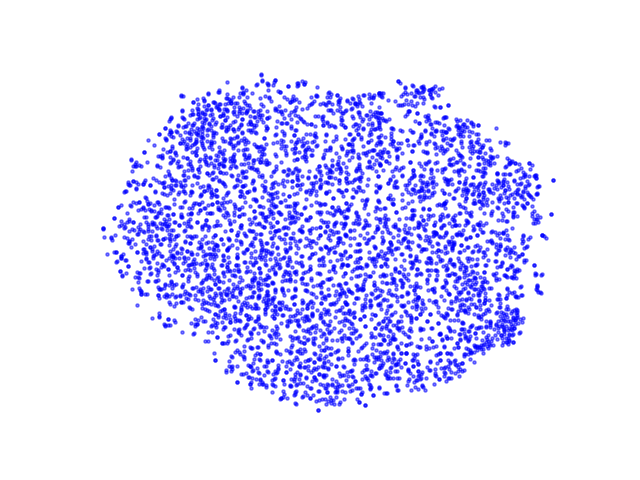}
    \caption{No Poison\newline\scriptsize}
    \label{fig:vis_none}
\end{subfigure}
\hfill
\centering
\begin{subfigure}{0.18\textwidth}
    \includegraphics[width=\textwidth]{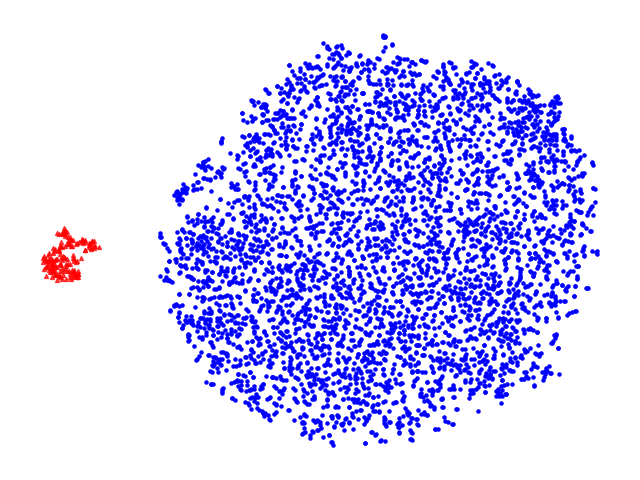}
     \caption{BadNet\newline\scriptsize\citet{gu2017badnets}}
     \label{fig:vis_badnet}
 \end{subfigure}
\hfill
\begin{subfigure}{0.18\textwidth}
    \includegraphics[width=\textwidth]{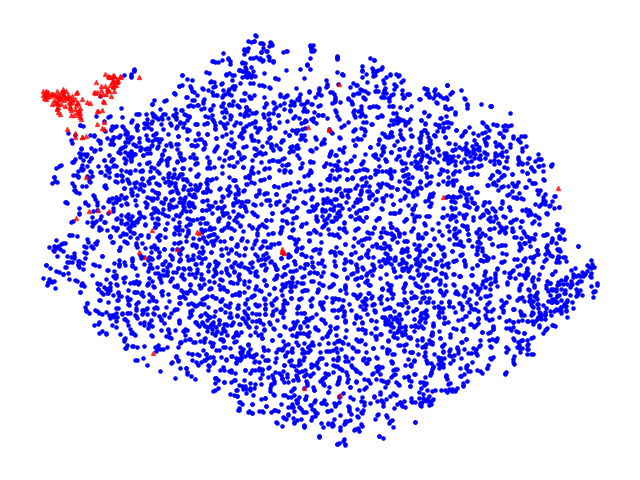}
    \caption{Blend\newline\scriptsize\citet{Chen2017TargetedBA}}
    \label{fig:vis_blend}
\end{subfigure}
\hfill
\begin{subfigure}{0.18\textwidth}
    \includegraphics[width=\textwidth]{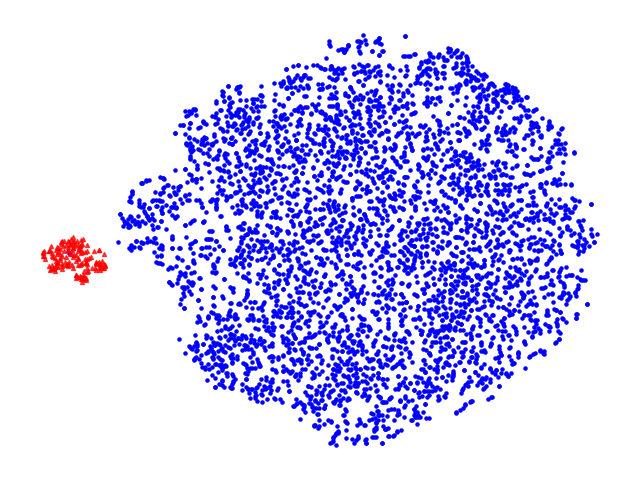}
    \caption{CL\newline\scriptsize\citet{turner2019label}}
    \label{fig:vis_clean_label}
\end{subfigure}
\hfill
%\begin{subfigure}{0.18\textwidth}
%    \includegraphics[width=\textwidth]{fig/tsne_all/tsne_cifar10_SIG_0.003_poison_seed=0_full_base_aug_seed=2333.pt_class=0.png}
%    \caption{SIG\newline\scriptsize\citet{barni2019new}}
%    \label{fig:SIG}
%\end{subfigure}
\begin{subfigure}{0.18\textwidth}
    \includegraphics[width=\textwidth]{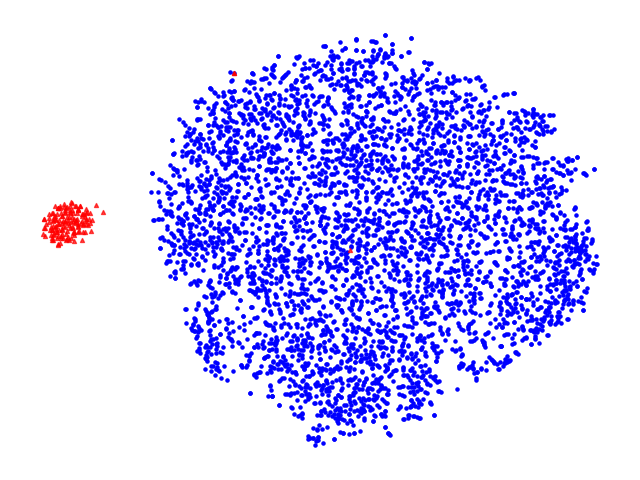}
    \caption{TaCT\newline\scriptsize\citet{tang2021demon}}
    \label{fig:vis_TaCT}
\end{subfigure}
\hfill
\\
\begin{subfigure}{0.18\textwidth}
    \includegraphics[width=\textwidth]{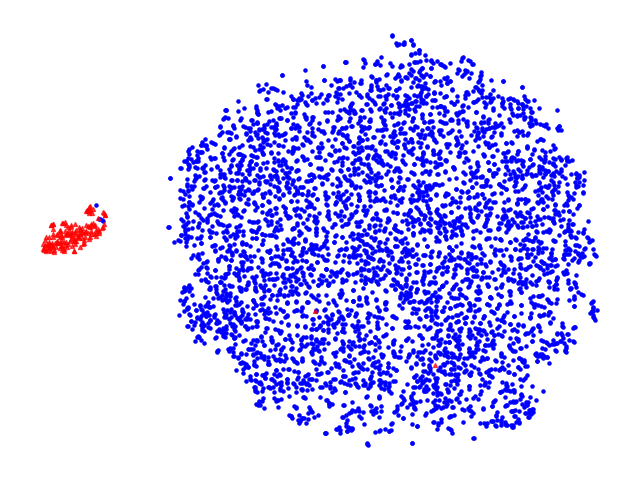}
    \caption{ISSBA\newline\scriptsize\citet{li2021invisible}}
    \label{fig:vis_ISSBA}
\end{subfigure}
\hfill
% \begin{subfigure}{0.18\textwidth}
%     \includegraphics[width=\textwidth]{fig/tsne_all/tsne_cifar10_WaNet_0.100_cover=0.200_poison_seed=0_full_base_aug_seed=2333.pt_class=0.png}
%     \caption{WaNet}
%     \label{fig:vis_WaNet}
% \end{subfigure}
% \hfill
\begin{subfigure}{0.18\textwidth}
    \includegraphics[width=\textwidth]{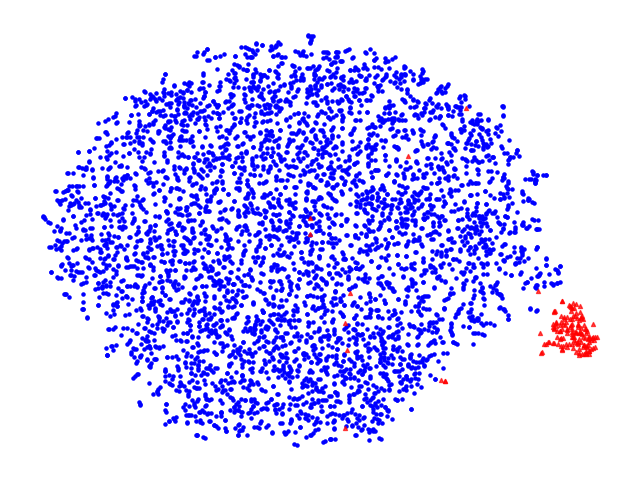}
    \caption{Dynamic\newline\scriptsize{\citet{nguyen2020input}}}
    \label{fig:vis_dynamic}
\end{subfigure}
\hfill
\begin{subfigure}{0.18\textwidth}
    \includegraphics[width=\textwidth]{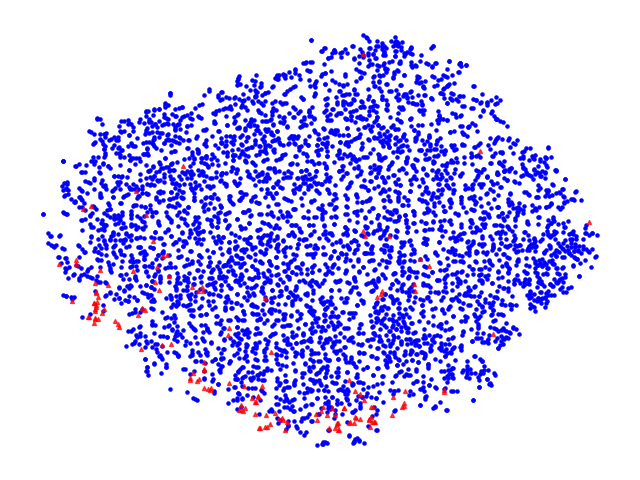}
    \caption{\scriptsize{\textbf{Adap-Blend\newline(Ours)}}}
    \label{fig:vis_adap_blend}
\end{subfigure}
\hfill
\begin{subfigure}{0.18\textwidth}
    \includegraphics[width=\textwidth]{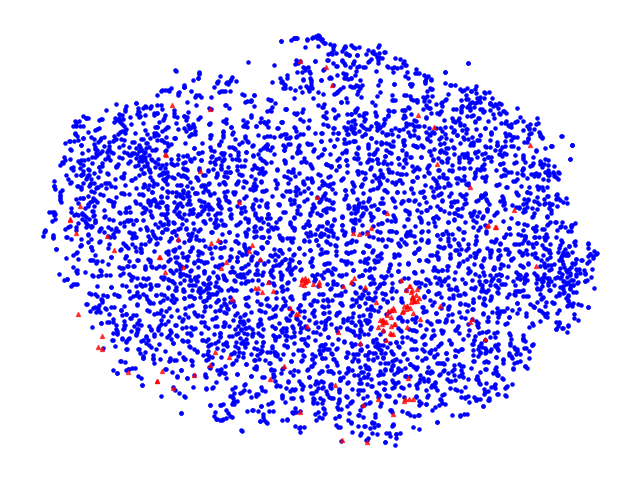}
    \caption{\scriptsize{\textbf{Adap-Patch\newline(Ours)}}}
    \label{fig:vis_adap_patch}
\end{subfigure}
        
\caption{T-SNE visualization of latent separability characteristic on CIFAR-10. Each point in the plots corresponds to a training sample from the target class. Caption of each subplot specifies its corresponding poison strategy. %All of the points are generated by projecting latent representations~(produced by the corresponding backdoored model for each poison strategy) of training samples to the top-2 PCA~\cite{pearson1901liii} plane. 
To highlight the separation, all \textcolor{red}{poison samples} are denoted by \textcolor{red}{red} points, while \textcolor{blue}{clean samples} correspond to \textcolor{blue}{blue} points. %Please refer Section~\ref{sec:why-these-defenses-fail} for more details.
} 
\label{fig:vis_separability}
\vspace{-0.6em}
\end{figure}

Despite the stealthiness in terms of model performance on natural samples, it has been commonly observed~\citep{tran2018spectral,chen2018activationclustering,huang2022backdoor} that backdoor poisoning attacks tend to leave tangible signatures in the latent space of backdoored models. As visualized in Fig~\ref{fig:vis_badnet} - Fig~\ref{fig:vis_dynamic}, poison and clean samples from the target class consistently form two separate clusters in the latent space, across a diverse set of backdoor poisoning attacks. %~\citep{gu2017badnets,Chen2017TargetedBA,turner2019label,tang2021demon}. 
The pervasiveness of the latent separation renders itself oftentimes as a default assumption, which we call \textit{latent separability assumption} in this work. %The pervasiveness of such phenomenons leads to the popularity of \textit{latent separability assumption}, which 
%Specifically, the assumption states that backdoored DNN models normally trained on a poisoned dataset via empirical risk minimization will unavoidably learn separable latent representations for poison and clean samples. 
A family of defenses~($i.e.$, \textit{latent separation based backdoor defenses}) explicitly base their designs on this assumption.  %From a defender's perspective, the latent separation characteristic provides a natural opportunity for designing backdoor defenses. 
These defenses first train a base classifier on the poisoned dataset, and expect the base model will naturally learn separable latent representations for poison and clean samples respectively. After that, they perform cluster analysis on the latent space of the base model. If the latent separation characteristics reliably arise, these defenses will be able to identify the outlier cluster formed by poison samples, and thus accurately filter out these poison samples from the training set. We note that \textit{this family of defenses are particularly important and successful in the backdoor defense literature}. Popular proposals in this family like Spectral Signature~\citep{tran2018spectral} and Activation Clustering~\citep{chen2018activationclustering} have already become indispensable baselines, and recent state-of-the-art proposals including SCAn~\citep{tang2021demon} and SPECTRE~\citep{hayase21a} in this family even claim to achieve nearly perfect recall with negligible false positive rate against a diverse set of attacks. Given the pervasiveness of the latent separation and its profound success in the application of backdoor defenses, a natural question arises: \textbf{\textit{Is the latent separation unavoidable for backdoor poisoning attacks?}}

\begin{figure}
\vspace{-2em}
\begin{center}
\includegraphics[width=\textwidth]{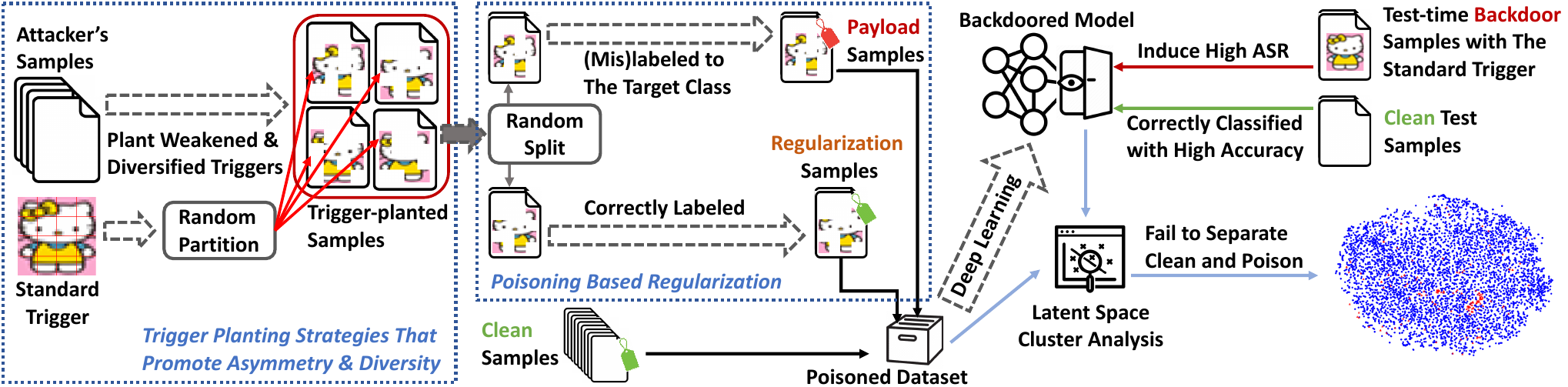}
\end{center}
\caption{
An overview of our adaptive backdoor poisoning attacks (here we take Adaptive-Blend introduced in Section ~\ref{subsec:adaptive_cases} as an example for illustration). Two key components render our attacks adaptive: \textbf{(1)} Poisoning-based regularization, which penalizes the backdoor correlation and helps to suppress the latent separation; \textbf{(2)} Trigger planting strategies that promote asymmetry and diversity, which help to maintain a high attack success rate as well as to improve latent space stealthiness. Please refer to our Section ~\ref{subsec:adaptive_with_cover} for more technical details.
}
\label{fig:overview}
\end{figure}

In this work, we revisit the assumption of latent separability and expose failure regions of defenses based on it. Specifically, we design adaptive backdoor {poisoning} attacks~(without control of the model training process), which can actively suppress the latent separation while maintaining a high attack success rate~(ASR) with negligible clean accuracy drop. %by sacrificing a controllable amount of attack success rate (ASR). On the one hand, 
%Such attacks have much weaker latent separation, and \textit{latent separation based backdoor defenses would thus be less effective~(and often completely fail)}. %On the other hand, although these attacks will sacrifice some ASR, they still keep considerably high ASR and will induce non-trivial damage to victim models. 
%In particular, we highlight two
%critical components that render our attacks adaptive~
Two critical components are underlying the design of our adaptive attacks (see Fig~\ref{fig:overview} for an overview): (1) \textit{Data poisoning based regularization}. After planting the backdoor trigger to a set of samples, we do not mislabel all of them to the target class. Instead, we randomly keep a fraction of them~(namely {regularization samples}) still correctly labeled to their real semantic classes. %~(other than the target class), and only label the rest~(namely {payload samples}) to the target class. 
Intuitively, these additional regularization samples penalize the backdoor correlation between the trigger and the target class. % --- fitting these samples will regularize the backdoor signal in the learned latent representations, and consequently the latent separation induced by the backdoor signal would also be suppressed. 
(2) \textit{Trigger planting strategies that promote asymmetry and diversity.} One may notice that penalization on the backdoor correlation induced by regularization samples can also greatly hurt the attack success rate~(ASR). We alleviate this problem via asymmetric trigger planting strategies. As illustrated in Fig~\ref{fig:overview}, we apply weakened triggers when we construct regularization and payload samples for data poisoning, while the original standard trigger would only be used during test time to activate the backdoor. %This design creates an asymmetry on backdoor triggers ---  weakened triggers for data poisoning and the (stronger) standard trigger for test-time attack.
Conceptually, in this way, since test-time backdoor samples (with the standard trigger) contain stronger backdoor features than those of regularization samples (with weakened triggers), the test-time attack can well mitigate the counter force from regularization samples and still maintain a high ASR. Besides asymmetry, our design also promotes diversity of triggers during data poisoning --- different poison samples could be stamped with different partial triggers, selected from a diverse set of trigger partitions. Intuitively, this diversity allows backdoor poison samples to scatter more diversely in the latent representation space, and can thus avoid being aggregated into an easy-to-identify cluster.

In conclusion, the main contributions of this paper are four-fold. \textbf{(1)} We confirm that the latent separability assumption holds across a diverse set of backdoor poisoning attacks in the existing literature. \textbf{(2)} We reveal that this assumption could fail, leading to poor performance of defenses that explicitly base their designs on it. \textbf{(3)} We design some simple yet effective adaptive backdoor poisoning attacks to present counter-examples against this assumption with two key novel components. \textbf{(4)} We conduct extensive experiments on benchmark datasets, verifying the effectiveness and the stealthiness in countering detection methods of our adaptive attacks.

\vspace{-2mm}

\section{Related Work}

\vspace{-2mm}
\paragraph{Backdoor Poisoning Attacks.} Backdoor poisoning attacks~\citep{gu2017badnets,Chen2017TargetedBA,turner2019label,li2021invisible} are also frequently referred to as poison-only backdoor attacks. This category of attacks only assume control over a small portion of the victim's training data, while the victim will train her own models on the poisoned dataset from scratch. Other backdoor attacks that assume additional control over the training process~\citep{shokri2020bypassing} or even weights of deployed models~\citep{liu2017fault,qi2021subnet,qi2022towards} do not fall in this category and are not considered in this work. %Different techniques for backdoor poisoning attacks have been extensively explored in recent years. \citet{turner2019label} propose clean label attacks, \citet{nguyen2020input,li2021invisible} explore sample specific triggers, while \citet{liu2020reflection,nguyen2021wanet} investigate backdoor triggers that are hard to be noticed by human inspectors.  
We refer readers to \citet{li2022backdoor} for a more comprehensive review.

\vspace{-3mm}
\paragraph{Latent Separation for Backdoor Defenses.}
%A commonly observed phenomenon following backdoor poisoning attacks is that --- 
It has been commonly observed~\citep{tran2018spectral} that models trained on a poisoned dataset tend to learn very different latent representations for backdoor and clean samples in the target class, which form two separate clusters~(see Fig~\ref{fig:vis_separability}). %This phenomenon is first identified by \citet{tran2018spectral}. They train a base backdoored model on the poisoned dataset and use it to map all training samples to the latent representation space. Then, for samples labeled to each class, they project their representations to the top PCA~\citep{pearson1901liii} direction. Interestingly, for all non-target classes that only contain clean samples, their latent representations regularly distribute around a single center; however, for the target class that contains both clean and poison samples, the latent representations consistently form a bimodal distribution --- poison and clean samples form their own modal respectively. 
This phenomenons is so pervasive that a family of defenses directly take the latent separation as a default assumption and propose to identify poison samples via performing cluster analysis on the latent space. This family includes Spectral Signature~\citep{tran2018spectral} and Activation Clustering~\citep{chen2018activationclustering}, which are most commonly evaluated baselines. More recent proposals~\citep{tang2021demon,hayase21a} in this family further claim to achieve nearly perfect recall with negligible false positive rate against a diverse set of attacks even in very low poison rate cases.

%and is then {utilized}~\citep{tran2018spectral,chen2018activationclustering,hayase21a,tang2021demon} to build backdoor defenses. Basically, one can perform cluster analysis on the latent representation space to detect bimodality and identify the separation boundary --- target class can thus be identified, and poison samples can also be eliminated by simply dropping all samples from the poison modal. %Later, \citet{chen2018activationclustering} also identify this phenomenon and propose to use clustering analysis to detect "poison cluster". More recent work like \citet{hayase21a} and \citet{tang2021demon} have further refined the detection performances by utilizing information of ``clean distribution''. 
%Although different techniques are applied, these defensive methods all \textbf{explicitly assume the separability} between poison and clean samples and indeed work quite well against many existing attacks --- in this work, we phrase the term \textbf{\underline{latent separability assumption}} to denote this common assumption.

\vspace{-3mm}
\paragraph{Adaptive Backdoor Attacks Against Latent Separation Based Defenses.} A family of adaptive backdoor attacks~\citep{tan2019bypassing,xia2021statistical,doan2021backdoor,ren2021simtrojan,cheng2020deep,zhong2022imperceptible} explicitly aim to reduce the latent separation between poison and clean samples. However, they do not fit into the paradigm of backdoor poisoning attacks --- they assume additional control over the whole training process and thus directly encode the latent inseparability into the training objectives of attacked models. A more relevant work is \citet{tang2021demon}, which points out that their source-specific poison-only attack can reduce latent separation. However, as shown in Fig~\ref{fig:vis_TaCT}, when the base model is trained along with standard data augmentation, \textit{there is still a notable separation} between the clean and poison populations, and actually \citet{tang2021demon} themselves also show that an improved latent space cluster analysis suffices to perfectly separate poison and clean samples of this attack. \textit{Thus, it is still unclear whether a poison-only backdoor attack can overcome the latent separation to evade backdoor defenses built on it.} In this work, we fill the gap and design adaptive backdoor poisoning attacks that can actively suppress the latent separation~(and thus circumvent existing latent separation based defenses).

%To our best knowledge, the first work that attempts to challenge the latent separation based backdoor defenses is \citet{shokri2020bypassing,xia2021statistical,doan2021backdoor,ren2021simtrojan,cheng2020deep,zhong2022imperceptible}, which has successfully bypassed existing backdoor detection algorithms by maximizing the indistinguishability between poison and clean samples in the latent space. \textit{However, this work assumes a stronger threat model} where adversaries not only control the training data but also control the whole training process --- thus they can directly encode the latent indistinguishability into the training objectives of the attacked models. Several more recent work~\citep{xia2021statistical,doan2021backdoor,ren2021simtrojan,cheng2020deep,zhong2022imperceptible} that also study this problem {all follow the same threat model} to \citet{shokri2020bypassing}. %For example, \citet{doan2021backdoor} directly adds a regularization term into the training loss to minimize the Wasserstein distance between poison and clean populations and \citet{xia2021statistical} adds a so-called Maximum Mean Discrepancy constraint to the loss function. 

\vspace{-3mm}
\paragraph{Other Backdoor Defenses.} %This work looks into \textit{adaptive} backdoor poisoning attacks \textit{target {specifically} to latent separation based defenses}, which reveals failure modes of the latent separability assumption. Note that, 
There are other defenses that are not built on latent separation. These include trigger synthesis~\citep{wang2019neural,AEVA}, model diagnosis~\citep{xu2019Meta,kolouri2020universal}, sample diagnosis~\citep{gao2019strip,guo2023scale}, fine-tuning~\citep{liu2017neural,li2021neural}, poison suppression~\citep{li2021anti,huang2022backdoor} and proactive training~\citep{qiproactive}, etc. Many of these proposals also have their own limitations revealed by existing literature~(refer \citet{li2022backdoor}), but they are not our focus in this work. %Still, for comprehensiveness, we reveal some of these defenses are also susceptible to our attacks in Appendix~\ref{appendix:more_defense_analysis}. \xiangyu{fix the link}

\vspace{-2mm}
\section{Notations and Threat Model}
\label{sec:problem_formulation}

\vspace{-3mm}
\paragraph{Notations.} We study image classification with DNN models. We denote a model by $\mathcal F_{\theta}: \mathcal X \mapsto [C]$, where $\theta$ are trainable parameters, $\mathcal X$ is the input space, $C$ is the number of classes, and $[C]:=\{1,2,\dots,C\}$. We decompose $\mathcal{F}_{\theta}$ as $\mathcal{F}_{\theta}=l_{\theta} \circ f_{\theta}$, where $l_\theta$ is the last linear prediction layer that transforms a latent representation into the final prediction label, and $f_\theta$ is the feature extractor. Given an input $x\in\mathcal X$, $f_{\theta}(x) \in \mathcal{H}$ is the latent representation of $x$ $w.r.t$ model $\mathcal{F}_{\theta}$, $\mathcal{H}$ denotes the latent representation space, and $\mathcal F_{\theta}(x)=l_{\theta}\circ f_{\theta}(x)$ is the predicted label. For backdoor poisoning attacks, we denote the clean training set by $\mathcal D = \{(x_i, y_i)\ |\ i=1,\dots,n\}$. We denote the backdoor trigger planting strategy by a transformation $\mathcal T: \mathcal X \mapsto \mathcal X$, and the adversary's poison label flipping strategy is denoted by $\mathcal L: \mathcal X \times [C] \mapsto [C]$. We use $\mathcal{J}:=\{j_1,\dots,j_p\}$ to denote indices of the $p$ data points that are controlled by the adversary. The resulting poisoned training set is denoted as $\mathcal{D}_{\text{poison}} = \{(\Tilde{x}_i, \Tilde{y}_i)\ |\ i = 1,\dots,n\}$, where
\vspace{-1mm}
\begin{align}
    & \Tilde{x}_i = 
    \begin{cases} \mathcal{T}(x_i) , & i \in \mathcal{J}\\
        x_i , & \text{otherwise}
    \end{cases},
    & \Tilde{y}_i = 
    \begin{cases} \mathcal{L}(x_i,y_i)  , & i \in \mathcal{J}\\
        y_i , & \text{otherwise}
    \end{cases}.\label{eqn:poison_formulation}
\end{align}
%Besides, $\rho = p/n$ is used to denote the poison rate and $t$ is the attacker-specified target class.

\vspace{-5mm}
\paragraph{Threat Model.}
\label{subsec:threat_model}

We consider the standard threat model of backdoor poisoning attacks~(poison-only backdoor attacks), where the adversary {\textit{only}} control a small portion of the victim's training data and the victim will \textit{train her own models from scratch} on the poisoned dataset manipulated by the adversary. %The adversary will modify the controlled training samples and turn the clean training set $D$ into a poisoned set $\mathcal{D}_{\text{poison}}$. By poisoning the victim's training dataset, the adversary aims at embedding backdoor into victim's models trained on the dataset. 
Specifically, the adversary will design a trigger planting strategy $\mathcal T$ and a label flipping strategy $\mathcal L$ to manipulate the controlled $p$ training samples~(as formulated in Eqn~\ref{eqn:poison_formulation}). A {victim model} %$\mathcal F_{\theta'}$ 
trained on 
%$\mathcal D_{\text{poison}}$ satisfies: $\mathbb P_{(x,y)\sim \mathbb B} [ \mathcal{F}_{\theta'}( \mathcal{T}(x) ) = t ] > \mathcal A$  and $\mathbb P_{(x,y)\sim \mathbb B} [ \mathcal{F}_{\theta'}(x) = y] \approx \mathbb P_{(x,y)\sim \mathbb B} [ \mathcal{F}_{\theta}(x) = y ]$, 
%\begin{align}
%\label{eq:asr}&\mathbb P_{(x,y)\sim \mathbb B}\Big( \mathcal{F}_{\theta'}( \mathcal{T}(x) ) = t\Big) \approx \mathcal A > \cfrac 1C\\
%\label{eq:ca}&\mathbb P_{(x,y)\sim \mathbb B}\Big( \mathcal{F}_{\theta'}(x) = y \Big) \approx \mathbb P_{(x,y)\sim \mathbb B}\Big( \mathcal{F}_{\theta}(x) = y \Big)
%\end{align}
%where $F_{\theta}$ is the {benign model} trained on the clean dataset $\mathcal{D}$.
the poisoned dataset $\mathcal{D}_{\text{poison}}$ will be backdoored --- that is, during test time, the model will (mis)classify a trigger-planted input to a target class $t$ with high probability, while keeping approximately the same performance to that of a benign model on genuine inputs. %Besides, the adversary may apply asymmetric triggers for data poisoning and test-time attack. That is, the adversary is allowed to choose a test-time trigger planting strategy $\mathcal T'$ different from $\mathcal{T}$. %enhanced test-time trigger planting strategy $\mathcal T'$ which plants stronger trigger pattern to the test-time input. 

\vspace{-2mm}
\section{Problem Formulation: Towards Poison-Only Backdoor attacks That Can Actively Suppress the Latent Separation}
\label{sec:latent_separability_formulation}

\vspace{-3mm}
\paragraph{Latent Separability Assumption for Backdoor Defense.} Given a poisoned dataset $\mathcal{D}_{\text{poison}}$, one can train a backdoored model $\mathcal F_{\theta}:=l_{\theta}\circ f_{\theta}$ via running a standard empirical risk minimization procedure $h$ on $\mathcal{D}_{\text{poison}}$, $i.e.$, $\theta \in  h(\mathcal{D}_{\text{poison}})$. Latent separability assumption indicates that, in the latent representation space generated by the backdoored model $\mathcal F_{\theta}$, poison and clean samples from the target class $t$ will form separate clusters, while samples from a non-target class only form a single homogeneous cluster~(see Fig~\ref{fig:vis_separability}). %Formally, we can consider an abstract heterogeneous criterion $\mathcal{I(\cdot,\cdot)}$ that takes two sets as input and generates a boolean output, indicating whether the two sets are heterogeneous~($i.e.$ form different clusters). Following the notations in Section~\ref{subsec:notations}, we use $H_B^c = \{f_{\theta'}(\Tilde{x}_i) | i \notin \mathcal{J} \land \Tilde{y}_i=c\}$ and $H_A^c = \{f_{\theta'}(\Tilde{x}_j)| j \in \mathcal{J} \land \Tilde{y}_j=c\}$ respectively to denote representations of clean samples and backdoor poison samples labeled as class $c$. Formally, {the latent separability assumption states that} there is a heterogeneous criterion $\mathcal{I}$ such that $\mathcal{I}(H_B^t, H_A^t) = True$ for target class $t$, and $\mathcal{I}(H_B^c, H_A^c) = False$ for any non-target class $c$. 
Latent separation based backdoor defenses~\citep{tran2018spectral,chen2018activationclustering,hayase21a,tang2021demon} propose to run cluster analysis on $H^c = \{f_{\theta}(\tilde{x}_i) | \tilde{y}_i=c\}$ for each class c. Typically, the defender will design a heterogeneous criterion $\mathcal{I(\cdot)}$ that takes $H^c$ as input and judges whether this set is heterogeneous~($i.e.$, contains separate clusters). On the heterogeneous $H^t$ identified by the criterion $\mathcal{I}$, the cluster analysis will divide $H^t$ into two empirical clusters ${H}_B^t$ and ${H}_A^t$, where ${H}_A^t$ is the suspected cluster formed by poison samples. The dataset will be cleansed by simply removing those training samples that generate ${H}_A^t$.

\vspace{-3mm}
\paragraph{Our Goals.} This work revisits the assumption of latent separability for backdoor defenses against poison-only backdoor attacks. We investigate \textit{adaptive backdoor poisoning attacks} that can actively suppress the latent separation between poison and clean samples. Ideally, against such adaptive attacks, the criterion $\mathcal{I}$ used by a defense should fail to detect the heterogeneity in $H^t$ and the cluster analysis would neither accurately separate poison and clean samples.

\vspace{-3mm}
\paragraph{Perspectives that Motivate Our Design.} %The latent separation induced by backdoor poisoning attacks has still not been well understood~(and are oftentimes overlooked), 
%Rigorously analyzing the latent separation induced by backdoor poisoning attacks is difficult in general. %On the one hand, the dynamics of training a DNN model on a poisoned dataset are extremely complicated. On the other hand, there is neither generally accepted definition for constraining patterns of backdoor triggers --- unlike the study on adversarial examples~\citep{szegedy2013intriguing,madry2018towards} that mainly focus on $\ell_p$-norm bounded perturbations, patterns of backdoor triggers used by existing attacks can be rather arbitrary~(e.g. patch~\citep{gu2017badnets}, image blending~\citep{Chen2017TargetedBA}, logo~\citep{liu2017trojaning}, sinusoidal signal~\citep{barni2019new}, natural reflection~\citep{liu2020reflection}, and even common objects~\citep{lin2020composite}). For these reasons, the latent separation characteristics have still not been well understood. This blank is very much concerning, because the latent separation 
%though it is among one of the earliest identified weaknesses~\citep{tran2018spectral} of backdoor poisoning attacks and it still arises so commonly in existing attacks~(including very recent ones). 
Two heuristic and mutually complementary perspectives on the latent separation phenomenon have inspired our design in this work. {The \underline{first} perspective} attributes the latent separation to \textit{the dominant impact of backdoor triggers}~\citep{tran2018spectral} during the inference of backdoored models. The intuition is --- in order to ``push'' a (trigger-planted) backdoor poison sample from its semantic class to the target class, a backdoored model tends to learn {an excessively strong signal for the backdoor trigger pattern} in latent representation space such that the signal can overwhelmingly beat other semantic features {to make its dictatorial decision}. The strong backdoor signal that exclusively appears in backdoor poison samples thus leads to the separation. The \underline{second} perspective is that, backdoored models learn separate representations for poison and clean samples simply because they tend to \textit{learn a separate shortcut rule}~\citep{geirhos2020shortcut} (solely based on the trigger pattern) to fit those poison samples without using any semantic features. The sense is --- backdoor learning is often independent of~(or only weakly correlated to) the semantic features used by the main task, thus the backdoored model that fits the poisoned dataset essentially just learns two unrelated~(or weakly related) tasks. From this aspect, there is not even an appealing reason for backdoor models to learn homogeneous latent representations for samples from the two heterogeneous tasks. Motivated by these perspectives, we conceive that a desirable adaptive backdoor poisoning attack (that can mitigate the latent separation) might need to encode some form of regularization, so as to (1) penalize the backdoored model for learning abnormally strong signals for the backdoor trigger; (2) encourage interconnection between backdoor learning and learning of the main task. These intuitions finally lead to our design in Section~\ref{sec:methodology}.

\vspace{-2mm}

\section{Our Methods}
\label{sec:methodology}

\vspace{-3mm}
We design {\textit{adaptive}} backdoor poisoning attacks following the insights we introduce in Section~\ref{sec:latent_separability_formulation}. %It turns out that a surprisingly simple adaptation suffices to implement those insights and can effectively suppress the latent separation. 
In Section~\ref{subsec:adaptive_with_cover}, we first present the generic framework underlying the design of our attacks. Then, in Section~\ref{subsec:adaptive_cases}, we elaborate concrete attacks that we implement in this work.

\vspace{-2mm}
\subsection{A Generic Framework for Adaptive Backdoor Poisoning Attacks}
\label{subsec:adaptive_with_cover}

\vspace{-2mm}

\paragraph{Overview.} We present an overview of our design in Fig~\ref{fig:overview}. Unlike typical backdoor poisoning attacks, in our framework, we do not label all trigger-planted samples to the target class. As shown, after planting the backdoor trigger to a set of samples~(sampled from all classes), we randomly split them into two disjoint groups. For one group, we still label them to the target class~(we call this group \textit{payload samples}) to establish the backdoor correlation between the trigger pattern and the target label;  while the other group~(namely \textit{regularization samples}) will instead be correctly labeled to their real semantic classes~(that can be diffident from the target class) to regularize  the backdoor correlation. Formally, following our notations in Section~\ref{sec:problem_formulation}, the adversary will specify a \textit{conservatism ratio} $\eta\in[0,1)$, with which our label flipping strategy formulates as:
\begin{align} \label{eq:label_strategy}
    \mathcal L(x_i,y_i) = 
    \begin{cases}
        t , & \text{with probability } 1-\eta\\
        y_i , & \text{with probability } \eta 
    \end{cases}.
\end{align}
Moreover, we introduce ideas of asymmetry and diversity into our trigger design --- we apply a diverse set of weakened triggers to construct regularization and payload samples for data poisoning, while the original standard trigger is used during test time to activate the backdoor. 
%When $\eta=0$, it degrades to a naive backdoor poisoning attack where all poison samples are labeled to the target class. %Formulation~\ref{eq:label_strategy} could be instantiated by specifying the \textit{regularization rate} and \textit{payload rate}, denoted as $\rho_r = \frac{\#regularization}{|\mathcal J|}$ and $\rho_p=\frac{\#payload}{|\mathcal J|}$ respectively, where $\frac{\rho_r}{\rho_r+\rho_p}=\eta$ and $\rho_r+\rho_p=\rho$. 
%Besides, the adversary uses asymmetric trigger planting strategies --- the weakened partial triggers $\mathcal{T}$ for data poisoning and the original full trigger $\mathcal{T}'$ for test-time attacks.

%Instead, we randomly keep a part~($\eta$) of them still correctly labeled to their semantic ground truth~(we call them \textit{regularization samples}), and only label the rest~(namely \textit{payload samples}) to the target class. Since these correctly labeled regularization samples weaken the backdoor correlation between the trigger and the target class, fitting these regularization samples will penalize the backdoor signal in the learned latent representations --- consequently, the latent separation induced by the backdoor signal should also be suppresse

\vspace{-0.8em}
\paragraph{Regularization Samples.} We note that, \textit{the introduction of regularization samples well incorporates our two insights from Section~\ref{sec:latent_separability_formulation}.} \underline{First}, with regularization samples, the backdoored model can no longer learn a dominantly strong signal for the backdoor trigger that dictatorially votes for the target class, otherwise, it can not fit regularization samples that are correctly labeled to other classes. This explains the naming of \textit{regularization samples} --- intuitively, they serve as regularizers that help to penalize the backdoor signal in the learned latent representations. \underline{Second}, the model can neither fit all trigger-planted samples via a simple shortcut rule. Instead, now \textit{it has to fit a much more complicated boundary} that should decide when to classify a trigger-planted input to the target class and when to classify it to its real semantic label, where the boundary is randomly generated. To successfully fit this boundary, the model must rely on both the trigger pattern and artifacts from the semantic features that coexist with the trigger, thus the learned latent representations for backdoor samples should be a more balanced fusion of both the trigger pattern and semantic features. 

\vspace{-0.8em}
\paragraph{Asymmetric Triggers.} \textit{The introduction of asymmetric triggers is critical for our attacks to still maintain a high attack success rate~(ASR).} As one may easily notice, since regularization samples penalize the backdoor correlation, a side-effect could be the drop of attack success rate~(ASR).  %Specifically, since we only mislabel $1-\eta$ of trigger-planted samples in the training set, models trained on the dataset are expected to capture a similar distribution and only misclassify $1-\eta$ test-time backdoor samples stamped with \textit{the same trigger}~(i.e. ASR drops approximately to $1-\eta$). 
To mitigate this problem, rather than using the same trigger for both data poisoning and test-time attack, in our design, we apply weakened triggers for data poisoning and use the (stronger) original {standard} trigger only for the test time. The intuition is: During test time, the backdoor samples (with the standard trigger) contain stronger backdoor features than those of regularization samples (with weakened triggers). This then enables test-time backdoor samples to have sufficient ``power" to mitigate the counter force from regularization samples and thus to still achieve a high ASR. We note that the idea of asymmetric triggers traces earliest back to \citet{Chen2017TargetedBA}, however the context is different. In order to evade human inspection on the poisoned dataset, \citet{Chen2017TargetedBA} propose to use weakened triggers that are visually less evident for data poisoning, and point out that a high ASR can still be maintained if the original standard trigger is used in test time. In our context, we use weakened triggers mainly to undermine the negative impact induced by regularization samples.

%The intuition is --- during test time, the full trigger now presents a stronger backdoor feature than the partial triggers in training-time poison samples, then the backdoor can thus be activated with a higher success rate, as earliest identified by \citet{Chen2017TargetedBA}. %The generic ideas can be summarized in two folds. First, adversaries can use weaker/partial triggers for data poisoning to evade dataset inspection \tinghao{talk about the heuristic here? maybe diversity?}, thus the backdoor can still be successfully embedded into the victim model. Second, during the test time, adversaries can still use the standard/full backdoor trigger to activate the embedded backdoor behaviors with high success rate.

%This design creates an asymmetry on backdoor triggers ---  weakened triggers for data poisoning and the (stronger) standard trigger for test-time attack. Conceptually, in this way, since test-time backdoor samples (with the standard trigger) contain stronger backdoor features than those of regularization samples (with weakened triggers), the test-time attack can well mitigate the counter force from regularization samples and still maintain a high ASR.

\vspace{-0.8em}
\paragraph{Trigger Diversification.} We also highlight that \textit{the trigger diversification in our design can also help our attacks to mitigate the latent separation.} Intuitively, since different poison samples could be planted with different triggers, these poison samples may scatter more diversely in the latent representation space. We expect such a more diverse scattering can prevent these poison samples from aggregating into an easy-to-identify cluster.

\vspace{-2mm}
\subsection{Instantiations of Our Attacks}
\label{subsec:adaptive_cases}

% \xiangyu{Introduce concrete instances that we implement, e.g. adaptive-k, adaptive-blend... What triggers, how to approach asymmetry?}

%\xiangyu{How to justify our choices? Why not xxx triggers? Why these triggers? Perhaps we need to provide a plausible explanation.}

%To instantiate concrete attacks within the framework we conceive, we need to further design suitable trigger planting strategies $\mathcal{T}$ and $\mathcal{T}'$ that would be used for data poisoning and test-time attacking. An interesting point revealed by Fig~\ref{fig:vis_separability} is that the simple Blend attack~(Fig~\ref{fig:vis_blend}) turns out to induce the least latent separation, better than many attacks that are usually deemed more advanced and stealthy. %, including clean label attack~(Fig~\ref{fig:vis_clean_labe}), source specific attack~(Fig~\ref{fig:vis_TaCT}) and sample-specific attack~(Fig~\ref{fig:vis_ISSBA}). 
%For this reason, we choose the Blend attack~\citep{Chen2017TargetedBA} as a start, and design \textit{Adaptive-Blend} via incorporating our adaptation techniques.

%An empirical observation from Fig~\ref{fig:vis_separability} is that the simple Blend attack~(Fig~\ref{fig:vis_blend}) turns out to induce the least latent separation, better than many attacks that are usually deemed more advanced and stealthy~(e.g. clean label attack~\ref{fig:vis_clean_labe}, source-specific attack~\ref{fig:vis_TaCT} and sample-specific attack~\ref{fig:vis_ISSBA}).

\vspace{-2mm}
Note that, our framework presented in Fig~\ref{fig:overview} is generic and can be creatively combined with existing techniques to instantiate powerful adaptive attacks. Following this framework, we instantiate two concrete attacks via directly adapting commonly used image blending based and patch based poison strategies, namely \textit{Adaptive-Blend} and \textit{Adaptive-Patch} respectively.

\vspace{-3mm}
\begin{wrapfigure}{l}{0.35\textwidth}

%\begin{figure}
\centering
\begin{subfigure}{0.15\textwidth}
    \includegraphics[width=\textwidth]{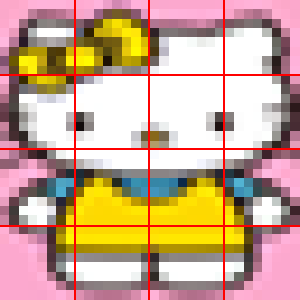}
    \label{fig:hellokitty_w_grids}
\end{subfigure}
\quad
\begin{subfigure}{0.15\textwidth}
    \includegraphics[width=\textwidth]{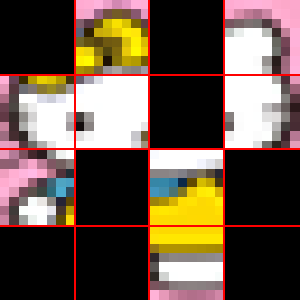}
    \label{fig:hellokitty_partitioned}
\end{subfigure}
\caption{In Adaptive-Blend, we partition the full trigger image into $4\times4=16$ pieces~(left), and randomly apply only $50\%$ of these trigger pieces~(right) to each poison sample, during data poisoning. Red lines demonstrate the grids by which we randomly mask the original trigger.}
\label{fig:adaptive_blend_instantiation}
%\end{figure}

\end{wrapfigure}

\paragraph{Adaptive-Blend.} An interesting point revealed by Fig~\ref{fig:vis_separability} is that the simple Blend attack~\citep{Chen2017TargetedBA} turns out to induce the least latent separation, better than many attacks that are usually deemed more advanced and stealthy. This suggests image blending based triggers as good candidates for designing attacks with weak latent separation. For this reason, we design \textit{Adaptive-Blend} via directly adapting the naive Blend attack according to our framework. Specifically, Adaptive-Blend introduces a conservatism ratio of $\eta = 0.5$ to balance the number of payload and regularization samples, and still adopts the process $\tilde x = (1-\alpha) \cdot x + \alpha \cdot T$ from \citet{Chen2017TargetedBA} to blend the trigger pattern $T$ into a genuine image $x$ to construct the trigger-planted sample $\tilde x$. As for asymmetric design, we still take the standard $\alpha = 0.2$ for test-time attacking, but use a weaker asymmetric opacity of $\alpha = 0.15$ for poison samples. Moreover, for a stronger and more diverse asymmetry, we propose to partition the standard full trigger~(Fig~\ref{fig:adaptive_blend_instantiation}, left) into $4\times 4=16$ pieces --- the full trigger would still be used for test-time attacking, while we randomly apply only $50\%$ of the partitioned trigger pieces~($e.g.$, Fig~\ref{fig:adaptive_blend_instantiation}, right) to each poison sample during data poisoning. This additional partition boosts both the ASR and the latent space stealthiness~(see Section~\ref{sec:ablation_necessary}).

\vspace{-3mm}
\paragraph{Adaptive-Patch.} Although the triggers could be more visually detectable, for comprehensiveness, we also instantiate our adaptive attack with patch based triggers, namely Adaptive-Patch. Empirically, since patch based triggers usually induce stronger latent separation~($e.g.$, see Fig~\ref{fig:vis_badnet}), we correspondingly turn to a larger conservatism ratio of $\eta = 2/3$. For trigger planting, rather than sticking to a single patch pattern, we prepare a more diverse set of 4 patch triggers (Fig~\ref{fig:poison-demo-adaptive-k-trigger-0}-\ref{fig:poison-demo-adaptive-k-trigger-3}) for data poisoning. Specifically, each poison sample is randomly attached to only one of the four triggers with a low opacity ($e.g.$, $50\%$) (Fig~\ref{fig:poison-demo-adaptive-k-poison-0}-\ref{fig:poison-demo-adaptive-k-poison-3}). At test time, we asymmetrically apply two (of the four) fully opaque triggers ($e.g.$, Fig~\ref{fig:poison-demo-adaptive-k-trigger-test}) simultaneously to achieve high ASR.

\vspace{-2mm}
\section{Experiments}
\label{sec:experiment}

% \xiangyu{Todos : 1. table-1 to compare existing attacks and our attacks in a certain poison rate; 2. Ablation on poison rate; 3. Reducing poison rate is not sufficient; 4. Asymmetric triggers are not sufficient; 5. Show the merits of asymmetry via an ablation on the cover rate --- ASR of symmetric attacks, ASR of asymmetric attacks, let's compare the two curves, etc.}

% \xiangyu{Perhaps, mention we also discuss some other non-latent separation based defenses in the appendix.}

\vspace{-1mm}
\subsection{Main Settings}
\label{subsec:experiment_setup}

\vspace{-2mm}
\paragraph{Datasets and Model Architectures.} We evaluate our adaptive attacks on three benchmark datasets that are commonly used in backdoor learning literature: CIFAR-10~\citep{krizhevsky2009learning}, GTSRB~\citep{stallkamp2012man} and a 10-classes subset of Imagenet~\citep{russakovsky2015imagenet}. For building base models, we also consider three different architectures including ResNet-20~\citep{he2016deep}, VGG-16~\citep{simonyan2014very} and Mobilenet-V2~\citep{sandler2018mobilenetv2}. Due to the space limit, in this section, we only present our results on CIFAR-10 with ResNet-20. We refer interested readers to Appendix~\ref{appendix:more_ablation_study} for results on other datasets and architectures. Detailed configurations on dataset split and training details of base models are deferred to Appendix~\ref{appendix:experiment_configurations}. 

%In Appendix~\ref{appendix:gtsrb}, we present the results for  GTSRB~\cite{stallkamp2012man}. In Appendix~\ref{appendix:other_architectures}, we also show qualitatively similar results for other model architectures. Refer Appendix~\ref{appendix:experiment_configurations}

\vspace{-3mm}
\paragraph{Attacks.} We evaluate our \texttt{Adap-Blend} and \texttt{Adap-Patch} attacks presented in Section~\ref{subsec:adaptive_cases}. We compare our adaptive attacks with six representative attacks in the literature. These attacks correspond to a diverse set of poisoning strategies including both classical and advanced ones. \texttt{BadNet}~\citep{gu2017badnets} and \texttt{Blend}~\citep{Chen2017TargetedBA} correspond to typical dirty-label attacks with patch-like triggers and blending based triggers respectively. \texttt{Dynamic}~\citep{nguyen2020input} and \texttt{ISSBA}~\citep{li2021invisible} correspond to input-aware backdoor attacks. \texttt{CL}~\citep{turner2019label} is a clean label attack. \texttt{TaCT}~\citep{tang2021demon} is a source-specific attack. Unless explicitly specified, for every attack, by default, we use 150 (payload) poison samples for data poisoning. Detailed attack configurations are described in Appendix~\ref{appendix:attack_configurations}.

\vspace{-3mm}
\paragraph{Defenses.} To validate the ``adaptiveness'' of our attacks against latent separation based backdoor defenses, we evaluate the four state-of-the-art defenses from this family: Spectral Signature~\citep{tran2018spectral}, Activation Clustering~\citep{chen2018activationclustering}, SCAn~\citep{tang2021demon} and SPECTRE~\citep{hayase21a}. All of these defenses are designed to detect and eliminate backdoor poison samples from the poisoned dataset, based on the assumed latent separation characteristics. 

% Note that, these defenses are exactly the main targets that our work aims to bypass. As a bonus, we also evaluate our attacks against some other typical backdoor defenses that do not explicitly rely on latent separability~\cite{liu2018fine,gao2019strip,wang2019neural,li2021anti}. Specifically, for STRIP~\cite{gao2019strip}, we consider both of its use cases, as a training set cleanser \textbf{STRIP(C)} and also as a test-time input filter \textbf{STRIP(F)}. Refer to Appendix~\ref{appendix:defense_configurations} for detailed configurations of all the defenses.

%We \underline{clean accuracy} and \underline{attack success rate} (ASR), the empirical probability that a backdoor input being classified to the target class. 
\vspace{-3mm}

\paragraph{Metrics.} For backdoor defenses that we evaluate, we measure their: 1) \textit{Elimination Rate}, ratio of (payload) poison samples that they successfully detect; 2) \textit{Sacrifice Rate}, ratio of clean samples falsely eliminated; 3) \textit{Attack Success Rate~(ASR)} of models retrained on the cleansed set; 4) \textit{Clean Accuracy} of models retrained on the cleansed set. Note that, ASR is defined as the ratio of trigger-planted samples that are mispredicted to the target class, while clean accuracy is the accuracy on genuine test samples. Moreover, to quantify the latent separation between clean and poison samples, we report the \textit{Silhouette Score}~\citep{rousseeuw1987silhouettes} of latent representations in the target class. A silhouette score is in the range from 0 to 1. A lower silhouette score indicates weaker separation. All the numbers that we report are average results across three independent repeated experiments.

% For (test-time) STRIP~\cite{gao2019strip}, we evaluate its effectiveness on 2,000 clean test samples and their 2,000 backdoor versions, by measuring the \textbf{elimination rate} and \textbf{sacrifice rate}. For Neural Cleanse~\cite{wang2019neural}, we report the \textbf{anomaly index} of the target class, and the unlearned model's \textbf{ASR} and \textbf{clean accuracy}. For Anti-Backdoor Learning (ABL~\cite{li2021anti}), we measure its \textbf{isolation precision}, \textbf{ASR} and \textbf{clean accuracy} of anti-backdoored models. To smooth the effect of randomness, we repeat all of our experiments for three times and report the average results. Finally, we notice that data augmentation sometimes benefits defenses, and therefore report the better defense result of the two models with and without augmentation.

% \subsection{Experiment Results}
% \label{subsec:experiment_results}

\begin{table}
\centering
\caption{Latent separability based defenses against our adaptive attacks on CIFAR-10.}
\vspace{-0.7em}
\resizebox{0.99\linewidth}{!}{ %< auto-adjusts font size to fill line
% \begin{tabular}{@{}lccc@{}}
\begin{tabular}{clccccccccc}
\toprule
 & ($\%$) & 
 \multicolumn{1}{c}{No Poison} &
\multicolumn{1}{c}{Blend} &
\multicolumn{1}{c}{BadNet} &
\multicolumn{1}{c}{ISSBA} &
\multicolumn{1}{c}{Dynamic} &
\multicolumn{1}{c}{CL} &
\multicolumn{1}{c}{TaCT} &
\multicolumn{1}{c}{\textbf{Adap-Blend (Ours)}} &
\multicolumn{1}{c}{\textbf{Adap-Patch (Ours)}}\cr
\cmidrule(lr){3-3} \cmidrule(lr){4-4}  \cmidrule(lr){5-5} \cmidrule(lr){6-6} \cmidrule(lr){7-7} \cmidrule(lr){8-8} \cmidrule(lr){9-9} \cmidrule(lr){10-10} \cmidrule(lr){11-11}

\multirow{2}*{\shortstack{Without Defense}}
& ASR & / & 89.0 & 99.9 & 95.3 & 97.5 & 93.6 & 96.5 & 76.5 & 97.5 \\
& Clean Accuracy & 92.0 & 91.7 & 91.5 & 91.6 & 91.8 & 92.1 & 91.8 & 91.6 & 91.5 \\
\midrule

% \multicolumn{8}{c}{\textbf{Defenses based on Latent Separability}} \cr
% \midrule

\multirow{4}*{\shortstack{Spectral\\Signature\\ \cite{tran2018spectral}}}
& Elimination Rate & / & 53.8 & 98.0 & 63.5 & 87.8 & 94.4 & 62.9 & 13.3 & 10.0 \cr
& Sacrifice Rate & 15.0 & 4.4 & 4.2 & 4.3 & 4.3 & 4.2 & 4.3 & 4.5 & 4.5  \cr
& ASR & / & \red{58.6} & 1.3 & 1.1 & \red{72.4} & \red{40.8} & \red{96.4} & \red{62.0} & \red{93.1} \\
& Clean Accuracy & 90.9 & 91.5 & 91.4 & 91.5 & 86.1 & 91.7 & 91.6 & 91.5 & 91.5 \\
\midrule

\multirow{4}*{\shortstack{Activation\\Clustering\\ \cite{chen2018activationclustering}}}
& Elimination Rate & / &  0.0 & 100.0 & 0.0 & 30.6 & 33.3 & 33.1 & 0.0 & 0.0 \cr
& Sacrifice Rate & 0.0 & 0.0 & 7.1 & 0.0 & 5.3 & 1.0 & 5.4 & 0.0 & 0.0 \cr
& ASR & / & \red{87.8} & 1.1 & \red{95.3} & \red{69.7} & \red{62.2} & \red{65.2} & \red{76.0} & \red{97.5} \\
& Clean Accuracy & 92.0 & 91.7 & 91.4 & 91.6 & 91.5 & 92.1 & 91.6 & 91.6 & 91.5 \\
\midrule

\multirow{4}*{\shortstack{SCAn\\ \cite{tang2021demon}}}
& Elimination Rate & / & 0.0 & 99.1 & 91.8 & 62.9 & 66.7 & 100.0 & 0.0 & 0.0 \cr
& Sacrifice Rate & 0.0 & 0.0 & 3.5 & 0.9 & 0.0 & 4.0 & 4.9 & 1.2 & 0.0 \cr
& ASR & / & \red{87.8} & 1.0 & 0.9 & \red{46.3} & \red{32.9} & 0.5 & \red{78.2} & \red{97.5} \\
& Clean Accuracy & 92.0 & 91.7 & 91.1 & 91.6 & 91.7 & 91.8 & 90.8 & 91.6 & 91.5 \\
\midrule

\multirow{4}*{\shortstack{SPECTRE\\ \cite{hayase21a}}}
& Elimination Rate & / & 96.4 & 100.0 & 100.0 & 99.8 & 100.0 & 100.0 & 6.9 & 0.0 \cr
& Sacrifice Rate & 1.5 & 0.2 & 0.2 & 0.2 & 0.2 & 0.2 & 0.2 & 0.5 & 0.5 \cr
& ASR & / & 5.7 & 0.8 & 1.0 & 7.7 & 1.6 & 1.7 & \red{69.0} & \red{94.8} \\
& Clean Accuracy & 91.6 & 91.7 & 91.7 & 91.6 & 91.6 & 91.6 & 91.6 & 91.4 & 91.6 \\
\midrule

\multicolumn{2}{c}{Silhouette Score} & / & 0.2608 & 0.4744 & 0.3933 & 0.4358 & 0.3964 & 0.2866 & \textbf{0.1065} & \textbf{0.0856} \\
\midrule

\end{tabular}
} % \resizebox

\label{tab:cifar10_main_results}
\end{table}

\vspace{-2mm}
\subsection{Main Results}
\label{subsec:attack_results}

\vspace{-1mm}

\begin{figure}
\centering
\begin{subfigure}{0.24\textwidth}
    \includegraphics[width=\textwidth]{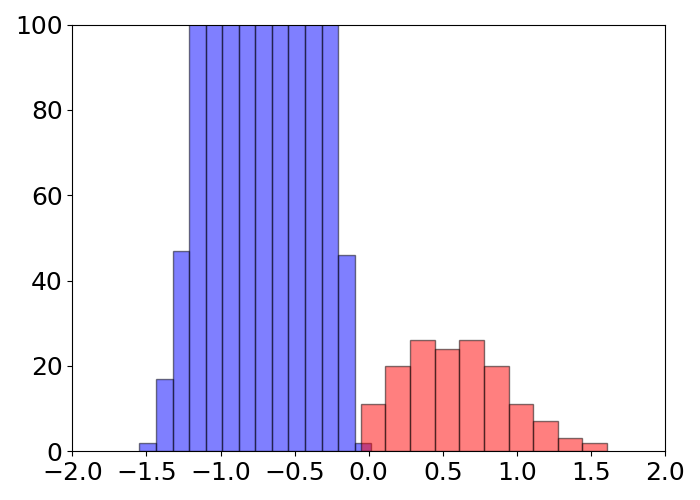}
    \caption{Blend}
    \label{fig:vis_svm_blend}
\end{subfigure}
\hfill
\begin{subfigure}{0.24\textwidth}
    \includegraphics[width=\textwidth]{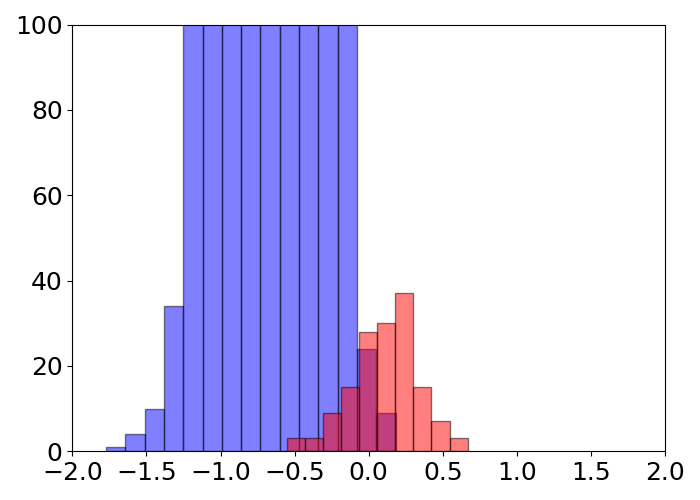}
    \caption{Adap-Blend}
    \label{fig:vis_svm_adap_blend}
\end{subfigure}
\hfill
\begin{subfigure}{0.24\textwidth}
    \includegraphics[width=\textwidth]{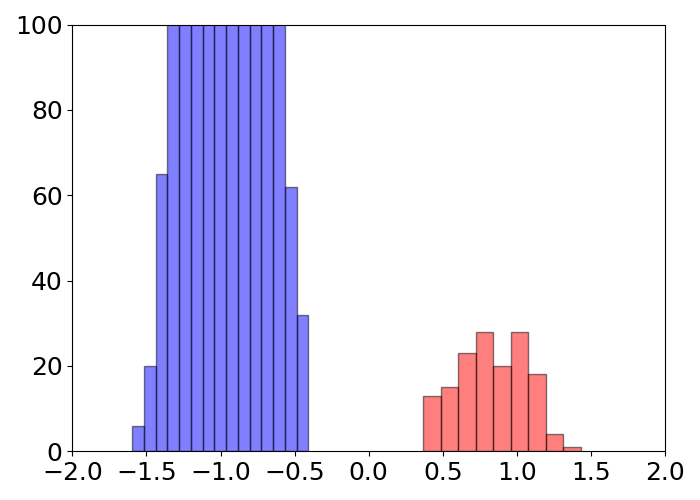}
    \caption{BadNet}
    \label{fig:vis_svm_badnet}
\end{subfigure}
\hfill
\begin{subfigure}{0.24\textwidth}
    \includegraphics[width=\textwidth]{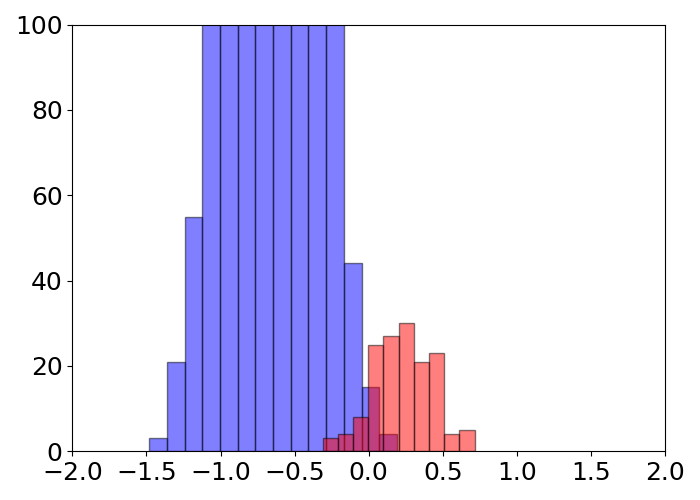}
    \caption{Adap-Patch}
    \label{fig:vis_svm_adap_patch}
\end{subfigure}
\vspace{-0.6em}
\caption{Visualization of latent representation spaces fitted by SVM. We use SVM to find the optimal (linear) boundary that separates poison and clean samples, and plot the histograms of (signed) distances between each point and the SVM hyperplane.}
\label{fig:vis_compare_svm}
\end{figure}

\paragraph{Visualization.} Fig~\ref{fig:vis_separability} plots latent representations of \textcolor{red}{poison} and \textcolor{blue}{clean} samples for different attacks, visualized by T-SNE~\citep{van2008visualizing}. As shown, notable latent separations are consistently observed on all the baseline attacks that we consider, while the poison and clean samples of our attacks mix with each other (Fig~\ref{fig:vis_adap_blend},\ref{fig:vis_adap_patch}). %Accordingly, we also quantify the latent separation by silhouette scores in the last row of Tab~\ref{tab:cifar10_main_results} --- our adaptive attacks have much smaller scores, indicating that the poison and clean samples are more overlapping and less separable. 
To further reveal the extent of latent inseparability of our adaptive attacks, we also use Support Vector Machine (SVM~\citep{cortes1995support}) to find the (linear) boundary that best separates poison and clean samples in the latent representation space. Fig~\ref{fig:vis_compare_svm} visualizes the histogram of distances between each data point and the SVM hyperplane. As shown, compared to non-adaptive attacks (Fig~\ref{fig:vis_svm_blend} and \ref{fig:vis_svm_badnet}), our adaptive attacks (Fig~\ref{fig:vis_svm_adap_blend} and \ref{fig:vis_svm_adap_patch}) bring the poison and clean samples much closer.

\vspace{-4mm}

\paragraph{Against Latent Separation Based Defenses.}

We present our main results in Table~\ref{tab:cifar10_main_results}. As shown, against SPECTRE~\citep{hayase21a} defense, the strongest latent separation based defense in the literature, none of the six baseline attacks survive --- SPECTRE can always eliminate almost all poison samples with negligible sacrifice of clean samples. This is consistent with our (qualitative) visualization in Fig~\ref{fig:vis_separability}, where notable latent separations are observed for all these attacks. It is also consistent with our quantitative measure of the latent separation --- all the six baseline attacks induce high Silhouette scores~($>0.25$). In comparison, our two adaptive attacks exhibit evident stealthiness in the latent representation space --- both the visualization results~(Fig~\ref{fig:vis_adap_blend},\ref{fig:vis_adap_patch}) and Silhouette scores indicate much weaker latent separation, and all the four latent separation based defenses are consistently \red{defeated}~(ASR is still larger than $20\%$ after defense). Besides, our adaptive attacks always achieve high ASR with negligible clean accuracy drop in all the cases.  When no defense is applied, both Adap-Blend ($>$75\%) and Adap-Patch achieve high ASR ($>$95\%). While none of the other six baseline attacks makes thorough all these defenses after cleansing and retraining, both our Adap-Blend and Adap-Patch consistently retain considerable ASR (Adap-Blend $>60\%$ and Adap-Patch $>90\%$), surviving each of them. We point out that, our results serve as counter examples against the assumption of latent separability and the ``adaptiveness" of our attacks are also validated. We also supplement results on other datasets and model architectures~(see  Appendix~\ref{appendix:more_ablation_study}).

%Refer Appendix~\ref{appendix:ablation_datasets} and \ref{appendix:other_architectures} for our results on other datasets and architectures.

%\paragraph{Circumventing Defenses constructed on Latent Separability.} 
%As shown in Table~\ref{tab:cifar10_main_results}, our adaptive strategy \red{successfully}~(ASR is still larger than $20\%$ after defense) evades all the four defenses built on the latent separability. Specifically, while none of the other seven poisoning backdoor attacks makes thorough all these defenses after cleansing and retraining, \textbf{both our Adap-Blend and Adap-Patch consistently retain considerable ASR (Adap-Blend $>60\%$ and Adap-Patch $>90\%$), surviving each of them}. Among these defenses, SPECTRE~\cite{hayase21a} performs as the strongest cleanser (eliminating almost every poison sample and introduce little false positive). However, against our adaptive attacks, SPECTRE either cannot detect the backdoor target class, or could only eliminate a tiny fraction of poison samples. %\tinghao{or just say SPECTRE fails?}

% \begin{figure}
% \centering
% \begin{subfigure}{0.45\textwidth}
%     \includegraphics[width=\textwidth]{fig/compare_svm/compare_svm_blend.png}
%     \caption{Blend~\cite{Chen2017TargetedBA}}
%     \label{fig:vis_compare_svm_blend}
% \end{subfigure}
% \quad
% \begin{subfigure}{0.45\textwidth}
%     \includegraphics[width=\textwidth]{fig/compare_svm/compare_svm_adaptive_blend.png}
%     \caption{Adaptive-Blend~(Ours)}
%     \label{fig:vis_compare_svm_adaptive_blend}
% \end{subfigure}
% \\
% \begin{subfigure}{0.45\textwidth}
% \includegraphics[width=\textwidth]{fig/compare_svm/compare_svm_k_triggers.png}
%     \caption{K Triggers}
%     \label{fig:vis_compare_svm_k_triggers}
% \end{subfigure}
% \quad
% \begin{subfigure}{0.45\textwidth}
%     \includegraphics[width=\textwidth]{fig/compare_svm/compare_svm_adaptive_k.png}
%     \caption{Adaptive-K~(Ours)}
%     \label{fig:vis_compare_svm_adaptive_k}
% \end{subfigure}
% \caption{\textbf{Visualization of latent spaces fit by SVM.}}
% \label{fig:vis_compare_svm}
% \end{figure}

\subsection{Ablation Studies}

%This subsection provides additional ablation studies to further supplement our results in Sec~\ref{subsec:attack_results}. %already show that our attack successfully bypass state-of-the-art backdoor defenses based on the latent separability assumption. Yet, it's still not clear why they fail to eliminate our adaptive poison and why our adaptive strategy works. In this section, 
%In particular, we look deeper and try to answer the following major questions: Q1) Why do latent separation based defenses fail against our adaptive attacks? Q2) How to understand regularization samples? Q3) Is every part of our adaptive attacks necessary? We provide our key observations that help to understand these questions.

\subsubsection{Poison Rate}
\label{subsec:ablation_poison_rate}

\begin{figure}
\centering
\begin{subfigure}{0.49\textwidth}
    \includegraphics[width=\textwidth]{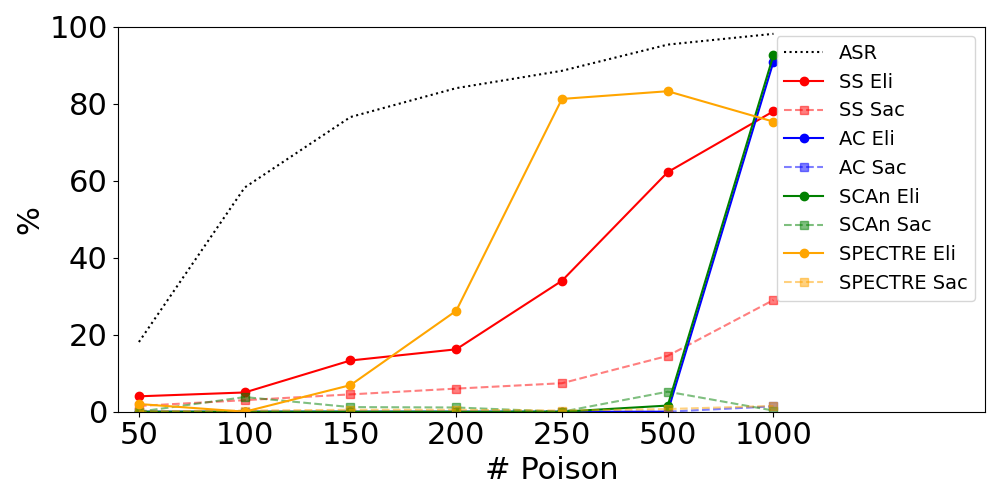}
    \caption{Adap-Blend}
    \label{fig:ablation_poison_rate_adap_blend}
\end{subfigure}
\hfill
\begin{subfigure}{0.49\textwidth}
    \includegraphics[width=\textwidth]{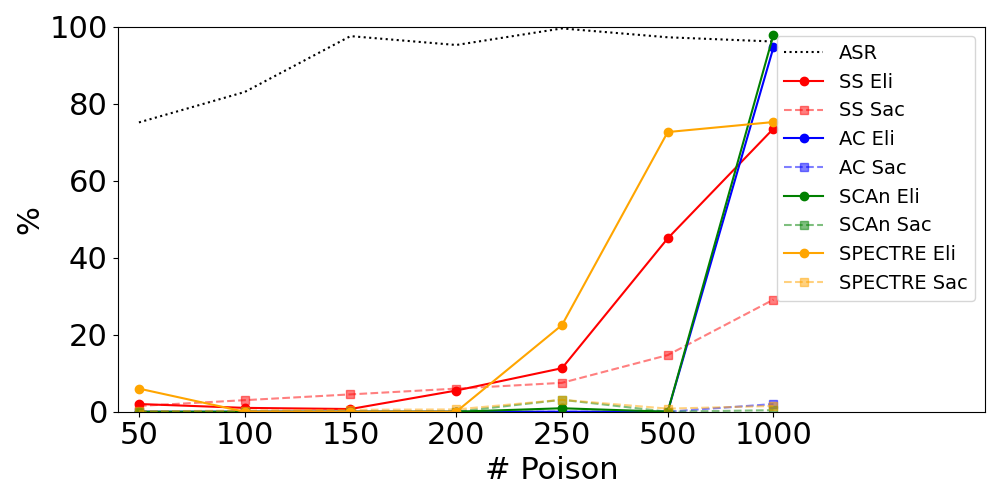}
    \caption{Adap-Patch}
    \label{fig:ablation_poison_rate_adap_patch}
\end{subfigure}
\vspace{-0.6em}
\caption{Defense results w.r.t. different (payload) poison samples. For Adap-Blend, we use as many regularization samples as payload samples; for Adap-Patch, we use twice the regularization sample number as payload samples. The black dotted lines show the ASR. We use different colors to represent the results of different defenses, where the solid lines correspond to Elimination (``Eli'') and the dotted lines correspond to Sacrifice (``Sac'').}
\label{fig:ablation_poison_rate}
\end{figure}

As mentioned in Section~\ref{subsec:experiment_setup}, our main experiments consistently use $150$ (payload) poison samples for poisoning attack. In Fig~\ref{fig:ablation_poison_rate}, we supplement additional results of our adaptive attacks with different number of poison samples. Specifically, we increase the number of payload poison samples from 50 to 1,000, and the number of regularization samples also proportionally vary according to the fixed conservatism ratio $\eta$ that we specify in Section~\ref{subsec:adaptive_cases}. As shown, one key takeaway is --- when the poison sample number grows too large ($e.g.$, 1,000), the stealthiness of our adaptive attacks start to significantly degrade. This is not surprising --- with more and more poison samples containing the rigid trigger pattern, the trigger pattern would become increasingly statistical significant, and models will unavoidably learn strong signal for this pattern in spite of the regularization. This indicates that a moderate poison rate is also a necessary condition for the success of our adaptive attacks.

\subsubsection{Strength of Regularization}

\begin{table}[!t]
\centering
\caption{Adap-Blend with different regularization sample numbers, with fixed 150 payload samples.}
\vspace{-0.7em}
\resizebox{0.8\linewidth}{!}{ %< auto-adjusts font size to fill line
\begin{tabular}{llcccccccccc}
\toprule
(\%) & \# Regularization Samples & 0 & 50 & 100 & 150 & 200 & 250 & 300 & 350 & 400 & 450 \cr
\cmidrule{3-3} \cmidrule{4-4} \cmidrule{5-5} \cmidrule{6-6} \cmidrule{7-7} \cmidrule{8-8} \cmidrule{9-9} \cmidrule{10-10} \cmidrule{11-11} \cmidrule{11-12}
& ASR & 89.0 & 86.5 & 83.9 & 76.5 & 74.1 & 70.4 & 65.6 & 60.9 & 58.0 & 56.5 \cr
\midrule
\multirow{3}*{\shortstack{SPECTRE}}
& Elimination Rate & 67.3 & 45.3 & 37.1 & 6.9 & 0.0 & 0.0 & 0.0 & 0.0 & 0.0 & 0.0  \cr
& Sacrifice Rate & 0.2 & 0.3 & 0.3 & 0.5 & 0.5 & 0.5 & 0.5 & 0.5 & 0.5 & 0.5  \cr
& ASR & 9.1 & 29.2 & 37.9 & 69.0 & 78.2 & 68.5 & 57.9 & 68.7 & 64.0 & 62.8 \cr
\bottomrule
\end{tabular}
} % \resizebox
\label{tab:ablation_regularization_samples}
\end{table}

Now, we fix the number of payload poison samples~(150 samples), and investigate varying number of regularization samples --- this reflects different strength of regularization. Specifically, we evaluate Adap-Blend against SPECTRE, and present the results in Tab~\ref{tab:ablation_regularization_samples}. We can generally tell that: 1) when the regularization is weak ($e.g.$, 0, 50, 100 regularization samples), our adaptive attacks could still be detected; 2) when the regularization is becoming stronger, our adaptive attacks start to mitigate the defense, though the ASR suffers from more sacrifice.

\subsubsection{Is every part of our adaptive strategy necessary?}
\label{sec:ablation_necessary}

\begin{table}[!t]
  \small
  \centering
  \caption{Ablation study to see if every part of our adaptive strategy is necessary.}
  \vspace{-0.4em}
  \subfloat[Blending attack with lower poison rates.\label{tab:ablation_blend_low_poison_rate}]{%
    \resizebox{0.46\linewidth}{!}{
    \begin{tabular}{lccc}
        \toprule
        (\%) & ASR & \multicolumn{2}{c}{SPECTRE}\cr
        \cmidrule{3-4}
        \# Poison Samples & & Elimination & Retrained ASR \cr
        \midrule
        50  & 69.7 & 84.0 & 8.7 \cr
        100 & 81.4 & 97.0 & 5.3 \cr
        150 & 89.0 & 96.4 & 5.7 \cr
        \bottomrule
    \end{tabular}%
    % \begin{tabular}{llccc}
    %     \toprule
    %     (\%) & \# Poison Samples & 50 & 100 & 150\cr
    %     \cmidrule{3-3}\cmidrule{4-4}\cmidrule{5-5}
    %     & ASR & 69.7 & 81.4 & 89.0 \cr
    %     \midrule
    %     \multirow{2}*{\shortstack{SPECTRE}}
    %     & Elimination Rate & 84.0 & 97.0 & 96.4 \cr
    %     & Sacrifice Rate & 0.1 & 0.1 & 0.2 \cr
    %     \bottomrule
    % \end{tabular}
    }
  }
  \hfill
  \subfloat[Adap-Blend with partial components.\label{tab:ablation_partial_component}]{%Adap-Blend simply using regularization samples (no asymmetric trigger) and simply using asymmetric trigger (no regularization samples) have larger silhouette score and fail to survive from SPECTRE, while our original Adap-Blend with both component survives
    \resizebox{0.5\linewidth}{!}{
    \begin{tabular}{lccc}
        \toprule
        (\%) & ASR & \multicolumn{2}{c}{SPECTRE}\cr
        \cmidrule{3-4}
        & & Elimination & Retrained ASR \cr
        \midrule
        No Diversity \& Asymmetry & 52.1 & 28.0 & 32.5 \cr
        No Regularization Samples & 89.0 & 67.3 & 9.1 \cr
        With Both & 76.5 & 6.9 & 69.0 \cr
        \bottomrule
    \end{tabular}%
    }
  }
\end{table}

% \begin{table}
% \centering
% \resizebox{0.4\linewidth}{!}{ %< auto-adjusts font size to fill line
% \begin{tabular}{llccc}
% \toprule
% Defense$\downarrow$ & \# Poison Samples $\rightarrow$ & 50 & 100 & 150\cr
% \midrule
% \multirow{2}*{\shortstack{SPECTRE}}
% & Elimination Rate & 84.0 & 97.0 & 96.4 \cr
% & Sacrifice Rate & 0.1 & 0.1 & 0.2 \cr
% \bottomrule
% \end{tabular}
% } % \resizebox
% \caption{Blending attack with lower poison rates.}
% \label{tab:ablation_blend_low_poison_rate}
% \end{table}

\paragraph{Simply reducing poison rate is not enough.} In Section~\ref{subsec:ablation_poison_rate}, we reveal that a low poison rate is necessary for the success of our attacks. Nonetheless, as shown in Tab~\ref{tab:ablation_blend_low_poison_rate}, when we lower the poison rate (to as few as 50 poison samples) of blending attack, it will still be cleansed by SPECTRE. Thus, simply reducing poison rate is not sufficient for mitigating the latent separation.

\vspace{-4mm}
\paragraph{Simply relying on regularization samples is not enough.} Regularization samples are important in our design, so is the trigger planting strategy we adopted (See Section~\ref{subsec:adaptive_with_cover} for discussion). If we use the standard symmetric trigger for both the data poisoning and test-time attack, both the ASR and latent space stealthiness would degrade --- the first row of Tab~\ref{tab:ablation_partial_component} shows Adap-Blend without asymmetric trigger partitioning, where it has a lower ASR ($\approx 50\%$) and could be further suppressed (retrained ASR $\approx 30\%$) by SPECTRE.

\vspace{-4mm}
\paragraph{Simply relying on asymmetric and diversified triggers is not enough.} Reversely, we study how our adaptive strategy behaves when we don't use regularization sample and rely solely on our trigger planting strategy. As shown in the second row of Tab~\ref{tab:ablation_partial_component}, though the original ASR gets higher, as a trade off, a much larger fraction ($>67\%$) of the poison samples can now be recognized and removed by SPECTRE and the retrained ASR drops severely. %We also launch a K-Patch attack, where we don't use regularization in the Adap-Patch setting. It turns out that 96.0\% poison samples of K-Patch could be eliminated by SPECTRE, and the ASR drops from 100\% to 2.8\% after retraining. 
This further confirms that regularization samples are vital for our attacks. See Fig~\ref{fig:ablation_no_reg} for a further $w.r.t.$ the necessity of regularization.

\vspace{-1mm}
\section{Discussions}
\label{sec:discussions}
\vspace{-1mm}
%\subsection{Negative Impacts}
%\label{subsec:negative_impacts}

%\subsection{Limitations}
%\label{subsec:limitations}

We expect an ideal adaptive attack can make the poison and clean samples completely indistinguishable in the latent space. This has been achieved under stronger threat model where the the training process is controlled~\citep{shokri2020bypassing,xia2021statistical,doan2021backdoor,ren2021simtrojan,cheng2020deep,zhong2022imperceptible}. In this paper, we take a step to this goal under poisoning-only threat model. We propose adaptive backdoor poisoning attacks that can suppress the latent separation and circumvent existing defenses based on it. However, as shown in Fig~\ref{fig:vis_compare_svm}, under oracle visualization, there is still a difference between poison and clean distributions, though the difference is greatly reduced. A remaining question is --- is it possible to achieve the ideal indistinguishable goals under the poison-only setting? We encourage future work to look into this question.

\section{Conclusion}

\vspace{-2mm}
In this work, we revisited the assumption of latent separability for backdoor defenses. We revealed that this assumption could fail, leading to the failure of backdoor defenses built on this assumption. Specifically, we provided our insights on the phenomenon of latent separation, and designed adaptive attacks that can mitigate this separation. Empirical study and evaluation of various latent separation-based defenses showed that our adaptive poisoning attacks indeed suppress the latent separation and render them ineffective. We call for defense designers to take caution when leveraging latent separability as an assumption in their defenses. We also encourage further defenses to take our attacks into consideration for a more comprehensive evaluation.

\section*{Ethics Statement}

During our study, we restricted all of our adversarial experiments within the laboratory environment, and did not induce any negative impact in the real world. Though our attacks may not be mitigated by many existing defenses, the illustration of our attacks is only conceptual. We encourage future work to design stronger defenses that resist our attacks.

\section*{Acknowledgements}

This work was supported in part by the National Science Foundation under grants CNS-1553437 and CNS-1704105, the ARL’s Army Artificial Intelligence Innovation Institute (A2I2), the Office of Naval Research Young Investigator Award, the Army Research Office Young Investigator Prize, Schmidt DataX award, Princeton E-ffiliates Award, and Princeton's Gordon Y. S. Wu Fellowship. Any opinions, findings, and conclusions or recommendations expressed in this material are those of the author(s) and do not necessarily reflect the views of the funding agencies.

\bibliography{iclr2023_conference}

\begin{thebibliography}{52}
\providecommand{\natexlab}[1]{#1}
\providecommand{\url}[1]{\texttt{#1}}
\expandafter\ifx\csname urlstyle\endcsname\relax
  \providecommand{\doi}[1]{doi: #1}\else
  \providecommand{\doi}{doi: \begingroup \urlstyle{rm}\Url}\fi

\bibitem[Barni et~al.(2019)Barni, Kallas, and Tondi]{barni2019new}
Mauro Barni, Kassem Kallas, and Benedetta Tondi.
\newblock A new backdoor attack in cnns by training set corruption without
  label poisoning.
\newblock In \emph{ICIP}, pp.\  101--105, 2019.

\bibitem[Borgnia et~al.(2021{\natexlab{a}})Borgnia, Cherepanova, Fowl, Ghiasi,
  Geiping, Goldblum, Goldstein, and Gupta]{borgnia2021strong}
Eitan Borgnia, Valeriia Cherepanova, Liam Fowl, Amin Ghiasi, Jonas Geiping,
  Micah Goldblum, Tom Goldstein, and Arjun Gupta.
\newblock Strong data augmentation sanitizes poisoning and backdoor attacks
  without an accuracy tradeoff.
\newblock In \emph{ICASSP}, pp.\  3855--3859, 2021{\natexlab{a}}.

\bibitem[Borgnia et~al.(2021{\natexlab{b}})Borgnia, Geiping, Cherepanova, Fowl,
  Gupta, Ghiasi, Huang, Goldblum, and Goldstein]{borgnia2021dp}
Eitan Borgnia, Jonas Geiping, Valeriia Cherepanova, Liam Fowl, Arjun Gupta,
  Amin Ghiasi, Furong Huang, Micah Goldblum, and Tom Goldstein.
\newblock Dp-instahide: Provably defusing poisoning and backdoor attacks with
  differentially private data augmentations.
\newblock In \emph{ICLR Workshop}, 2021{\natexlab{b}}.

\bibitem[Chen et~al.(2019)Chen, Carvalho, Baracaldo, Ludwig, Edwards, Lee,
  Molloy, and Srivastava]{chen2018activationclustering}
Bryant Chen, Wilka Carvalho, Nathalie Baracaldo, Heiko Ludwig, Benjamin
  Edwards, Taesung Lee, Ian Molloy, and Biplav Srivastava.
\newblock Detecting backdoor attacks on deep neural networks by activation
  clustering.
\newblock In \emph{AAAI Workshop}, 2019.

\bibitem[Chen et~al.(2017)Chen, Liu, Li, Lu, and Song]{Chen2017TargetedBA}
Xinyun Chen, Chang Liu, Bo~Li, Kimberly Lu, and Dawn Song.
\newblock Targeted backdoor attacks on deep learning systems using data
  poisoning.
\newblock \emph{arXiv preprint arXiv:1712.05526}, 2017.

\bibitem[Cheng et~al.(2021)Cheng, Liu, Ma, and Zhang]{cheng2020deep}
Siyuan Cheng, Yingqi Liu, Shiqing Ma, and Xiangyu Zhang.
\newblock Deep feature space trojan attack of neural networks by controlled
  detoxification.
\newblock In \emph{AAAI}, 2021.

\bibitem[Cortes \& Vapnik(1995)Cortes and Vapnik]{cortes1995support}
Corinna Cortes and Vladimir Vapnik.
\newblock Support-vector networks.
\newblock \emph{Machine learning}, 20\penalty0 (3):\penalty0 273--297, 1995.

\bibitem[Doan et~al.(2021)Doan, Lao, and Li]{doan2021backdoor}
Khoa Doan, Yingjie Lao, and Ping Li.
\newblock Backdoor attack with imperceptible input and latent modification.
\newblock In \emph{NeurIPS}, 2021.

\bibitem[Fastai(2019)]{imagenette}
Fastai.
\newblock Fastai/imagenette: A smaller subset of 10 easily classified classes
  from imagenet, and a little more french, 2019.
\newblock URL \url{https://github.com/fastai/imagenette}.

\bibitem[Gao et~al.(2019)Gao, Xu, Wang, Chen, Ranasinghe, and
  Nepal]{gao2019strip}
Yansong Gao, Change Xu, Derui Wang, Shiping Chen, Damith~C Ranasinghe, and
  Surya Nepal.
\newblock Strip: A defence against trojan attacks on deep neural networks.
\newblock In \emph{ACSAC}, 2019.

\bibitem[Geirhos et~al.(2020)Geirhos, Jacobsen, Michaelis, Zemel, Brendel,
  Bethge, and Wichmann]{geirhos2020shortcut}
Robert Geirhos, J{\"o}rn-Henrik Jacobsen, Claudio Michaelis, Richard Zemel,
  Wieland Brendel, Matthias Bethge, and Felix~A Wichmann.
\newblock Shortcut learning in deep neural networks.
\newblock \emph{Nature Machine Intelligence}, 2\penalty0 (11):\penalty0
  665--673, 2020.

\bibitem[Gu et~al.(2017)Gu, Dolan-Gavitt, and Garg]{gu2017badnets}
Tianyu Gu, Brendan Dolan-Gavitt, and Siddharth Garg.
\newblock Badnets: Identifying vulnerabilities in the machine learning model
  supply chain.
\newblock \emph{arXiv preprint arXiv:1708.06733}, 2017.

\bibitem[Guo et~al.(2022)Guo, Li, and Liu]{AEVA}
Junfeng Guo, Ang Li, and Cong Liu.
\newblock {AEVA}: Black-box backdoor detection using adversarial extreme value
  analysis.
\newblock In \emph{ICLR}, 2022.

\bibitem[Guo et~al.(2023)Guo, Li, Chen, Guo, Sun, and Liu]{guo2023scale}
Junfeng Guo, Yiming Li, Xun Chen, Hanqing Guo, Lichao Sun, and Cong Liu.
\newblock Scale-up: An efficient black-box input-level backdoor detection via
  analyzing scaled prediction consistency.
\newblock In \emph{ICLR}, 2023.

\bibitem[Hayase et~al.(2021)Hayase, Kong, Somani, and Oh]{hayase21a}
Jonathan Hayase, Weihao Kong, Raghav Somani, and Sewoong Oh.
\newblock Spectre: defending against backdoor attacks using robust statistics.
\newblock In \emph{ICML}, pp.\  4129--4139, 2021.

\bibitem[He et~al.(2016)He, Zhang, Ren, and Sun]{he2016deep}
Kaiming He, Xiangyu Zhang, Shaoqing Ren, and Jian Sun.
\newblock Deep residual learning for image recognition.
\newblock In \emph{CVPR}, pp.\  770--778, 2016.

\bibitem[Huang et~al.(2022)Huang, Li, Wu, Qin, and Ren]{huang2022backdoor}
Kunzhe Huang, Yiming Li, Baoyuan Wu, Zhan Qin, and Kui Ren.
\newblock Backdoor defense via decoupling the training process.
\newblock In \emph{ICLR}, 2022.

\bibitem[Kolouri et~al.(2020)Kolouri, Saha, Pirsiavash, and
  Hoffmann]{kolouri2020universal}
Soheil Kolouri, Aniruddha Saha, Hamed Pirsiavash, and Heiko Hoffmann.
\newblock Universal litmus patterns: Revealing backdoor attacks in cnns.
\newblock In \emph{CVPR}, pp.\  301--310, 2020.

\bibitem[Krizhevsky(2012)]{krizhevsky2009learning}
Alex Krizhevsky.
\newblock Learning multiple layers of features from tiny images.
\newblock \emph{University of Toronto}, 05 2012.

\bibitem[Li et~al.(2021{\natexlab{a}})Li, Lyu, Koren, Lyu, Li, and
  Ma]{li2021anti}
Yige Li, Xixiang Lyu, Nodens Koren, Lingjuan Lyu, Bo~Li, and Xingjun Ma.
\newblock Anti-backdoor learning: Training clean models on poisoned data.
\newblock In \emph{NeurIPS}, 2021{\natexlab{a}}.

\bibitem[Li et~al.(2021{\natexlab{b}})Li, Lyu, Koren, Lyu, Li, and
  Ma]{li2021neural}
Yige Li, Xixiang Lyu, Nodens Koren, Lingjuan Lyu, Bo~Li, and Xingjun Ma.
\newblock Neural attention distillation: Erasing backdoor triggers from deep
  neural networks.
\newblock In \emph{ICLR}, 2021{\natexlab{b}}.

\bibitem[Li et~al.(2021{\natexlab{c}})Li, Zhai, Jiang, Li, and
  Xia]{li2021backdoor}
Yiming Li, Tongqing Zhai, Yong Jiang, Zhifeng Li, and Shu-Tao Xia.
\newblock Backdoor attack in the physical world.
\newblock In \emph{ICLR Workshop}, 2021{\natexlab{c}}.

\bibitem[Li et~al.(2022)Li, Jiang, Li, and Xia]{li2022backdoor}
Yiming Li, Yong Jiang, Zhifeng Li, and Shu-Tao Xia.
\newblock Backdoor learning: A survey.
\newblock \emph{IEEE Transactions on Neural Networks and Learning Systems},
  2022.

\bibitem[Li et~al.(2021{\natexlab{d}})Li, Li, Wu, Li, He, and
  Lyu]{li2021invisible}
Yuezun Li, Yiming Li, Baoyuan Wu, Longkang Li, Ran He, and Siwei Lyu.
\newblock Invisible backdoor attack with sample-specific triggers.
\newblock In \emph{ICCV}, pp.\  16463--16472, 2021{\natexlab{d}}.

\bibitem[Liu et~al.(2018)Liu, Dolan-Gavitt, and Garg]{liu2018fine}
Kang Liu, Brendan Dolan-Gavitt, and Siddharth Garg.
\newblock Fine-pruning: Defending against backdooring attacks on deep neural
  networks.
\newblock In \emph{RAID}, pp.\  273--294, 2018.

\bibitem[Liu et~al.(2017{\natexlab{a}})Liu, Wei, Luo, and Xu]{liu2017fault}
Yannan Liu, Lingxiao Wei, Bo~Luo, and Qiang Xu.
\newblock Fault injection attack on deep neural network.
\newblock In \emph{ICCAD}, pp.\  131--138, 2017{\natexlab{a}}.

\bibitem[Liu et~al.(2020)Liu, Ma, Bailey, and Lu]{liu2020reflection}
Yunfei Liu, Xingjun Ma, James Bailey, and Feng Lu.
\newblock Reflection backdoor: A natural backdoor attack on deep neural
  networks.
\newblock In \emph{ECCV}, pp.\  182--199, 2020.

\bibitem[Liu et~al.(2017{\natexlab{b}})Liu, Xie, and Srivastava]{liu2017neural}
Yuntao Liu, Yang Xie, and Ankur Srivastava.
\newblock Neural trojans.
\newblock In \emph{ICCD}, pp.\  45--48, 2017{\natexlab{b}}.

\bibitem[Nguyen \& Tran(2021)Nguyen and Tran]{nguyen2021wanet}
Anh Nguyen and Anh Tran.
\newblock Wanet--imperceptible warping-based backdoor attack.
\newblock In \emph{ICLR}, 2021.

\bibitem[Nguyen \& Tran(2020)Nguyen and Tran]{nguyen2020input}
Tuan~Anh Nguyen and Anh Tran.
\newblock Input-aware dynamic backdoor attack.
\newblock In \emph{NeurIPS}, pp.\  3454--3464, 2020.

\bibitem[Peng et~al.(2022)Peng, Xiong, Sun, and Li]{peng2022label}
Minlong Peng, Zidi Xiong, Mingming Sun, and Ping Li.
\newblock Label-smoothed backdoor attack.
\newblock \emph{arXiv preprint arXiv:2202.11203}, 2022.

\bibitem[Qi et~al.(2021)Qi, Zhu, Xie, and Yang]{qi2021subnet}
Xiangyu Qi, Jifeng Zhu, Chulin Xie, and Yong Yang.
\newblock Subnet replacement: Deployment-stage backdoor attack against deep
  neural networks in gray-box setting.
\newblock \emph{arXiv preprint arXiv:2107.07240}, 2021.

\bibitem[Qi et~al.(2022{\natexlab{a}})Qi, Xie, Pan, Zhu, Yang, and
  Bu]{qi2022towards}
Xiangyu Qi, Tinghao Xie, Ruizhe Pan, Jifeng Zhu, Yong Yang, and Kai Bu.
\newblock Towards practical deployment-stage backdoor attack on deep neural
  networks.
\newblock In \emph{CVPR}, pp.\  13347--13357, 2022{\natexlab{a}}.

\bibitem[Qi et~al.(2022{\natexlab{b}})Qi, Xie, Wang, Wu, Mahloujifar, and
  Mittal]{qiproactive}
Xiangyu Qi, Tinghao Xie, Jiachen~T. Wang, Tong Wu, Saeed Mahloujifar, and
  Prateek Mittal.
\newblock Towards a proactive ml approach for detecting backdoor poison
  samples, 2022{\natexlab{b}}.
\newblock URL \url{https://arxiv.org/abs/2205.13616}.

\bibitem[Ren et~al.(2021)Ren, Li, and Zhou]{ren2021simtrojan}
Yankun Ren, Longfei Li, and Jun Zhou.
\newblock Simtrojan: Stealthy backdoor attack.
\newblock In \emph{ICIP}, pp.\  819--823, 2021.

\bibitem[Rousseeuw(1987)]{rousseeuw1987silhouettes}
Peter~J Rousseeuw.
\newblock Silhouettes: a graphical aid to the interpretation and validation of
  cluster analysis.
\newblock \emph{Journal of computational and applied mathematics}, 20:\penalty0
  53--65, 1987.

\bibitem[Russakovsky et~al.(2015)Russakovsky, Deng, Su, Krause, Satheesh, Ma,
  Huang, Karpathy, Khosla, Bernstein, et~al.]{russakovsky2015imagenet}
Olga Russakovsky, Jia Deng, Hao Su, Jonathan Krause, Sanjeev Satheesh, Sean Ma,
  Zhiheng Huang, Andrej Karpathy, Aditya Khosla, Michael Bernstein, et~al.
\newblock Imagenet large scale visual recognition challenge.
\newblock \emph{International journal of computer vision}, 115\penalty0
  (3):\penalty0 211--252, 2015.

\bibitem[Sandler et~al.(2018)Sandler, Howard, Zhu, Zhmoginov, and
  Chen]{sandler2018mobilenetv2}
Mark Sandler, Andrew Howard, Menglong Zhu, Andrey Zhmoginov, and Liang-Chieh
  Chen.
\newblock Mobilenetv2: Inverted residuals and linear bottlenecks.
\newblock In \emph{CVPR}, pp.\  4510--4520, 2018.

\bibitem[Shen et~al.(2021)Shen, Ji, Zhang, Li, Chen, Shi, Fang, Yin, and
  Wang]{shen2021backdoor}
Lujia Shen, Shouling Ji, Xuhong Zhang, Jinfeng Li, Jing Chen, Jie Shi,
  Chengfang Fang, Jianwei Yin, and Ting Wang.
\newblock Backdoor pre-trained models can transfer to all.
\newblock In \emph{CCS}, 2021.

\bibitem[Shokri et~al.(2020)]{shokri2020bypassing}
Reza Shokri et~al.
\newblock Bypassing backdoor detection algorithms in deep learning.
\newblock In \emph{EuroS\&P}, pp.\  175--183, 2020.

\bibitem[Simonyan \& Zisserman(2014)Simonyan and Zisserman]{simonyan2014very}
Karen Simonyan and Andrew Zisserman.
\newblock Very deep convolutional networks for large-scale image recognition.
\newblock \emph{arXiv preprint arXiv:1409.1556}, 2014.

\bibitem[Stallkamp et~al.(2012)Stallkamp, Schlipsing, Salmen, and
  Igel]{stallkamp2012man}
Johannes Stallkamp, Marc Schlipsing, Jan Salmen, and Christian Igel.
\newblock Man vs. computer: Benchmarking machine learning algorithms for
  traffic sign recognition.
\newblock \emph{Neural networks}, 32:\penalty0 323--332, 2012.

\bibitem[Tan \& Shokri(2020)Tan and Shokri]{tan2019bypassing}
Te~Juin~Lester Tan and Reza Shokri.
\newblock Bypassing backdoor detection algorithms in deep learning.
\newblock In \emph{EuroS\&P}, 2020.

\bibitem[Tang et~al.(2021)Tang, Wang, Tang, and Zhang]{tang2021demon}
Di~Tang, XiaoFeng Wang, Haixu Tang, and Kehuan Zhang.
\newblock Demon in the variant: Statistical analysis of dnns for robust
  backdoor contamination detection.
\newblock In \emph{USENIX Security}, 2021.

\bibitem[Tran et~al.(2018)Tran, Li, and Madry]{tran2018spectral}
Brandon Tran, Jerry Li, and Aleksander Madry.
\newblock Spectral signatures in backdoor attacks.
\newblock \emph{arXiv preprint arXiv:1811.00636}, 2018.

\bibitem[Turner et~al.(2019)Turner, Tsipras, and Madry]{turner2019label}
Alexander Turner, Dimitris Tsipras, and Aleksander Madry.
\newblock Label-consistent backdoor attacks.
\newblock \emph{arXiv preprint arXiv:1912.02771}, 2019.

\bibitem[Van~der Maaten \& Hinton(2008)Van~der Maaten and
  Hinton]{van2008visualizing}
Laurens Van~der Maaten and Geoffrey Hinton.
\newblock Visualizing data using t-sne.
\newblock \emph{Journal of machine learning research}, 9\penalty0 (11), 2008.

\bibitem[Wang et~al.(2019)Wang, Yao, Shan, Li, Viswanath, Zheng, and
  Zhao]{wang2019neural}
Bolun Wang, Yuanshun Yao, Shawn Shan, Huiying Li, Bimal Viswanath, Haitao
  Zheng, and Ben~Y Zhao.
\newblock Neural cleanse: Identifying and mitigating backdoor attacks in neural
  networks.
\newblock In \emph{IEEE S\&P}, pp.\  707--723, 2019.

\bibitem[Xia et~al.(2022)Xia, Niu, Li, and Li]{xia2021statistical}
Pengfei Xia, Hongjing Niu, Ziqiang Li, and Bin Li.
\newblock Enhancing backdoor attacks with multi-level mmd regularization.
\newblock \emph{IEEE Transactions on Dependable and Secure Computing}, 2022.

\bibitem[Xie et~al.(2019)Xie, Huang, Chen, and Li]{xie2019dba}
Chulin Xie, Keli Huang, Pin-Yu Chen, and Bo~Li.
\newblock {DBA}: Distributed backdoor attacks against federated learning.
\newblock In \emph{ICLR}, 2019.

\bibitem[Xu et~al.(2021)Xu, Wang, Li, Borisov, Gunter, and Li]{xu2019Meta}
Xiaojun Xu, Qi~Wang, Huichen Li, Nikita Borisov, Carl~A Gunter, and Bo~Li.
\newblock Detecting ai trojans using meta neural analysis.
\newblock In \emph{IEEE S\&P}, 2021.

\bibitem[Zhong et~al.(2022)Zhong, Qian, and Zhang]{zhong2022imperceptible}
Nan Zhong, Zhenxing Qian, and Xinpeng Zhang.
\newblock Imperceptible backdoor attack: From input space to feature
  representation.
\newblock In \emph{IJCAI}, 2022.

\end{thebibliography}
\bibliographystyle{iclr2023_conference}

\newpage

\appendix

\section{Experiment Configurations}\label{appendix:experiment_configurations}

\subsection{Computational Environments}
\label{appendix:computational_environments}

All of our experiments are conducted on a workstation with 48 Intel Xeon Silver 4214 CPU cores, 384 GB RAM, and 8 GeForce RTX 2080 Ti GPUs.  

\subsection{Attack Configurations}
\label{appendix:attack_configurations}

\begin{figure}[tbp]
	\centering
	\subfloat[]{\label{fig:poison-demo-blend-trigger}\includegraphics[width=0.65in]{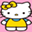}} \quad
	\subfloat[]{\label{fig:poison-demo-adaptive-blend-trigger}\includegraphics[width=0.65in]{fig/poison_demo/hellokitty_partitioned.png}} \quad
	\subfloat[]{\label{fig:poison-demo-adaptive-k-trigger-0}\includegraphics[width=0.65in]{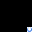}} \quad
	\subfloat[]{\label{fig:poison-demo-adaptive-k-trigger-1}\includegraphics[width=0.65in]{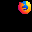}} \quad
	\subfloat[]{\label{fig:poison-demo-adaptive-k-trigger-2}\includegraphics[width=0.65in]{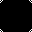}} \quad
	\subfloat[]{\label{fig:poison-demo-adaptive-k-trigger-3}\includegraphics[width=0.65in]{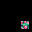}} \quad
	\subfloat[]{\label{fig:poison-demo-adaptive-k-trigger-test}\includegraphics[width=0.65in]{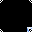}} \quad
% 	\subfloat[Clean]{\label{fig:poison-demo-clean}\includegraphics[width=0.65in]{fig/poison_demo/demo_clean.png}} \quad
	\\
	\subfloat[\scriptsize{Adap-Blend (Test) \& Blend}]{\label{fig:poison-demo-blend-poison}\includegraphics[width=0.65in]{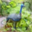}} \quad
	\subfloat[\scriptsize{Adap-Blend}]{\label{fig:poison-demo-adaptive-blend-poison}\includegraphics[width=0.65in]{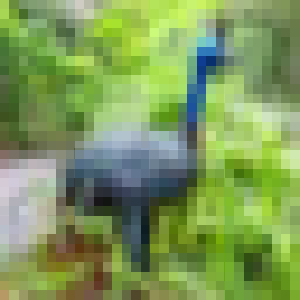}} \quad
	\subfloat[\scriptsize{1$^\text{st}$ Trigger}]{\label{fig:poison-demo-adaptive-k-poison-0}\includegraphics[width=0.65in]{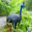}} \quad
	\subfloat[\scriptsize{2$^\text{nd}$ Trigger}]{\label{fig:poison-demo-adaptive-k-poison-1}\includegraphics[width=0.65in]{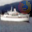}} \quad
	\subfloat[\scriptsize{3$^\text{rd}$ Trigger}]{\label{fig:poison-demo-adaptive-k-poison-2}\includegraphics[width=0.65in]{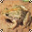}} \quad
	\subfloat[\scriptsize{4$^\text{th}$ Trigger}]{\label{fig:poison-demo-adaptive-k-poison-3}\includegraphics[width=0.65in]{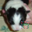}} \quad
    \subfloat[\scriptsize{Adap-K (Test)}]{\label{fig:poison-demo-adaptive-k-poison-test}\includegraphics[width=0.65in]{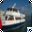}}
	\caption{The example of poisoned samples. } \label{fig:poison-demo}
\end{figure}

For each attack, we reuse the same triggers adopted in the paper. ``Blend'' and ``Adap-Blend'' use the blending-trigger (Fig~\ref{fig:poison-demo-blend-trigger}) with 20\% opacity. ``TaCT'' uses the trojan square trigger (Fig\ref{fig:poison-demo-adaptive-k-poison-3}) with 100\% opacity. Eventually, ``Adap-Patch'' use the four triggers in Fig~\ref{fig:poison-demo-adaptive-k-trigger-0}-\ref{fig:poison-demo-adaptive-k-trigger-3} with 50\%, 20\%, 50\% and 30\% opacities, respectively. Specifically, each of their poison training sample randomly selects one of the four triggers (see Fig~\ref{fig:poison-demo-adaptive-k-poison-0}-\ref{fig:poison-demo-adaptive-k-poison-3}), while each test sample uses the enhanced two triggers demonstrated in Fig~\ref{fig:poison-demo-adaptive-k-poison-test}.

% \paragraph{Fairness Concern of Trigger Selections.} For other attacks, we follow the trigger selections adopted in their original works in order to objectively evaluate how they perform against the defenses. For our attack, we argue that \textbf{our trigger design itself is a vital part of our design} --- our triggers should not necessarily be the same to triggers of other attacks. Still, when we design Adap-Blend, we use \textbf{the same trigger} from Blend. The comparison between Blend and Adapt-Blend fairly reflects the effectiveness of our adaptive strategies.

The target class is set to class 0. 150 poison samples are used for ``Blend'', ``BadNet'', ``ISSBA'', ``Dynamic'' and ``CL''.  For ``TaCT'', we use 150 poison samples and 150 cover samples. For ``Adap-Blend'', we use 150 payload samples and 150 regularization samples. For our ``Adap-Patch'', we use 150 payload samples and 300 regularization samples.

% \paragraph{Fairness Concern of Poison Rates.} In our work, we do not consider the (correctly labeled) regularization samples malicious, since they could not inject backdoor independently. Only payload poison samples mislabelled to the target class are considered to be malicious. On the other hand, the stealthiness in the latent space is also strongly correlated with the number of payload samples --- more malicious payload samples empirically lead to less stealthiness in the latent representation space, and thus are easier to be defended. So in Table~\ref{tab:cifar10_main_results} of our paper, all attacks use the same number of malicious payload poison samples (exactly 150) \textbf{in the target class}, meeting the criteria of fairness in the sense of stealthiness.

For all backdoor models, we adopt the standard training pipeline. SGD with a momentum of 0.9, a weight decay of $10^{-4}$, and a batch size of 128, is used for optimization. Initially, we set the learning rate to $0.1$. On CIFAR-10, we follow the standard 200 epochs stochastic gradient descent procedure, and the learning rate will be multiplied by a factor of $0.1$ at the epochs of $100$ and $150$. On GTSRB we use 100 epochs of training, and the learning rate is multiplied by $0.1$ at the epochs of $40$ and $80$.

We consider widely adopted data augmentations, i.e. \texttt{RandomHorizontalFlip} and \texttt{RandomCrop} for CIFAR-10 and \texttt{RandomRotation} for GTSRB. We notice that data augmentation may affect defense results significantly, while defenders do not know whether to use augmentation or not. Therefore, we report the better defense result of the two backdoor models with and without augmentation. In another sentence, we report \textbf{upper-bound results} of the defenses which might be affected by the incorporation of data augmentation during backdoor training. Also, to rule out the effect of randomness, we train three models on three seeds for each configuration and report their average results.

\subsubsection{Fairness Considerations}

\paragraph{Trigger Selections.} For other non-adaptive attacks, we follow the trigger selections adopted in their original works in order to objectively evaluate how they perform against the defenses. For our attack, we argue that our trigger design itself is a vital part of our design --- our triggers should not necessarily be the same to triggers of other attacks. Still, when we design Adap-Blend, we use the same trigger from Blend. The comparison between Blend and Adapt-Blend fairly reflects the effectiveness of our adaptive strategies. As for Adaptive-Patch, since it's difficult to partition the BadNet trigger further, instead, we turn to another set of patch triggers, which might be a potential unfair factor. To further address this fairness concern, hereby we provide in Tab~\ref{tab:ablation_k_trigger} an extra set of controlled experiment (the poison rate follows the configurations used for Tab~\ref{tab:cifar10_main_results} in our paper): \texttt{K-trigger} attack that uses the same set of triggers as Adap-Patch for poisoning and testing, only without our adaptive regularization. Like Blend vs. Adap-Blend, K-trigger can thus be deemed as a non-adaptive counterpart to Adaptive-patch. As shown, K-trigger can still be eliminated by most of the latent-space defenses.

\begin{table}[!t]
\centering
\caption{Additional results of comparing K-trigger and Adap-Patch attacks. Configurations are similar to Tab~\ref{tab:cifar10_main_results}.}
\vspace{-0.7em}
\resizebox{0.99\linewidth}{!}{ %< auto-adjusts font size to fill line
% \begin{tabular}{@{}lccc@{}}
\begin{tabular}{ccccccccccccccc}
\toprule
\textbf{Defenses$\rightarrow$} &
\multicolumn{2}{c}{No Defense} &
\multicolumn{2}{c}{Spectral Signature} &
\multicolumn{2}{c}{Activation Clustering} & 
\multicolumn{2}{c}{SCAn} & 
\multicolumn{2}{c}{SPECTRE} \cr
\cmidrule(lr){2-3} \cmidrule(lr){4-5} \cmidrule(lr){6-7} \cmidrule(lr){8-9} \cmidrule(lr){10-11} \cmidrule(lr){12-13}
\textbf{Attacks} $\downarrow$ & ASR & CA & Elimination & Sacrifice & Elimination & Sacrifice & Elimination & Sacrifice & Elimination & Sacrifice \cr
\midrule
K-trigger & 100.0 & 91.5 & 62.7 & 4.3 & 0.0 & 3.5 & 92.0 & 11.5 & 94.7 & 0.2 \cr
\midrule
Adap-Patch & 98.8 & 91.0 & 10.0 & 4.5 & 0.0 & 0.0 & 0.0 & 0.0 & 0.0 & 0.5 \cr
\bottomrule

\end{tabular}
} % \resizebox
\label{tab:ablation_k_trigger}
\end{table}

\paragraph{Poison Rates.} We do not consider the (correctly labeled) regularization samples ``malicious in our context'', since they could not inject backdoor independently and do not induce latent separation. Only payload poison samples mislabelled to the target class are considered to be ``malicious in our context''. On the other hand, the stealthiness in the latent space is also strongly correlated with the number of payload samples --- more malicious payload samples empirically lead to less stealthiness in the latent representation space, and thus are easier to be defended. So in Table~\ref{tab:cifar10_main_results}, all attacks use the same number of malicious payload poison samples (exactly 150) in the target class, meeting the criteria of fairness in the sense of stealthiness. But, indeed, since Adap-Blend also uses an additional 150 (non-malicious) poison samples for regularization, it uses 300 poison samples in total, more than that of other non-adaptive attacks --- but if we also use 300 poison samples for those non-adaptive attacks, they would be even less stealthy in the latent representation space, which is also unfair.

Still, we may want to know, \textit{what if the total number of poison samples (counting both payload and regularization samples) are set the same?} To address this concern, in Tab~\ref{tab:ablation_300_samples}, we supplement an extra set of experiment results on non-adaptive attacks, where each attack inject exactly 300 poison samples in total (same as Adap-Blend). Obviously, all these non-adaptive attacks still fail against latent-space defenses.

\begin{table}[!t]
\centering
\caption{Additional results of non-adaptive attacks with 300 poison samples in total. Configurations are similar to Tab~\ref{tab:cifar10_main_results}.}
\vspace{-0.7em}
\resizebox{0.99\linewidth}{!}{ %< auto-adjusts font size to fill line
% \begin{tabular}{@{}lccc@{}}
\begin{tabular}{ccccccccccccccc}
\toprule
\textbf{Defenses$\rightarrow$} &
\multicolumn{2}{c}{No Defense} &
\multicolumn{2}{c}{Spectral Signature} &
\multicolumn{2}{c}{Activation Clustering} & 
\multicolumn{2}{c}{SCAn} & 
\multicolumn{2}{c}{SPECTRE} \cr
\cmidrule(lr){2-3} \cmidrule(lr){4-5} \cmidrule(lr){6-7} \cmidrule(lr){8-9} \cmidrule(lr){10-11} \cmidrule(lr){12-13}
\textbf{Attacks} $\downarrow$ & ASR & CA & Elimination & Sacrifice & Elimination & Sacrifice & Elimination & Sacrifice & Elimination & Sacrifice \cr
\midrule
Blend & 95.3 & 91.8 & 67.0 & 8.6 & 85.3 & 16.1 & 94.7 & 0.0 & 98.7 & 0.3 \cr
\midrule
BadNet & 100.0 & 91.8 & 100.0 & 8.5 & 100.0 & 0.0 & 100.0 & 0.0 & 100.0 & 0.3 \cr
\midrule
ISSBA & 98.7 & 91.7 & 19.3 & 8.9 & 0.0 & 0.0 & 97.0 & 0.0 & 99.7 & 0.3 \cr
\midrule
Dynamic & 98.9 & 91.5 & 72.3 & 8.6 & 95.7 & 3.1 & 96.3 & 0.0 & 100.0 & 0.3 \cr
\midrule
CL & 99.6 & 92.0 & 100.0 & 8.5 & 100.0 & 0.1 & 100.0 & 0.0 & 100.0 & 0.3 \cr
\midrule
TaCT & 96.5 & 91.8 & 62.9 & 4.3 & 33.1 & 5.4 & 100.0 & 4.9 & 100.0 & 0.2 \cr
\bottomrule

\end{tabular}
} % \resizebox
\label{tab:ablation_300_samples}
\end{table}

\subsection{Defense Configurations}
\label{appendix:defense_configurations}

\begin{itemize}
    \item Spectral Signature~\citep{tran2018spectral} removes $1.5 \cdot \rho_p$ suspected samples from every class.
    \item Activation Clustering~\citep{chen2018activationclustering} removes clusters with size $<$35\% of the class size.
    \item SCAn~\citep{tang2021demon} cleanses classes with scores larger than $e$.
    \item SPECTRE~\citep{hayase21a} removes $1.5 \cdot \rho_p$ suspected samples only from the class with the highest QUE score.
\end{itemize}

\section{More Ablation Study}
\label{appendix:more_ablation_study}

Since GTSRB and Imagenette contain different number of samples. For simplicity, we use the ratio of the whole training set to denote the number of poison samples, $i.e.$, poison rate. We use $\rho$ to denote the total ratio of poison samples. For our adaptive attacks, we use $\rho_p$ to denote the ratio of payload poison samples and $\rho_c$ to denote the ratio of regularization samples. Basically, $\rho = \rho_p + \rho_c$.

\subsection{Adaptive Attacks on Different Datasets}
\label{appendix:ablation_datasets}

In this subsection, we evaluate our attacks on different datasets. The same ResNet-20 architecture that we use in the main experiment is consistently used for all the experiments in this subsection. 

\subsubsection{GTSRB}
\label{appendix:gtsrb}

GTSRB~\citep{stallkamp2012man} is a widely adopted dataset for backdoor study, for which we also evaluate our adaptive poisoning backdoor attacks on GTSRB.

Due to the imbalanced nature of GTSRB and the rotation-based data augmentation, we activate Adap-Patch with another set of asymmetric triggers, $i.e.$, Fig~\ref{fig:poison-demo-adaptive-k-trigger-1} and Fig~\ref{fig:poison-demo-adaptive-k-trigger-3}, for high ASR. We use $\rho_p=0.003$ and $\rho_c=0.003$ for ``Adap-Blend''. We notice that ``Adap-Patch'' has a lower ASR on GTSRB than on CIFAR-10 (due to more classes); to boost up the ASR, we use $\rho_p=0.005$ and $\rho_c=0.01$ for ``Adap-Patch''. For comparison, we also show results of ``BadNet'' ($\rho=0.005$), ``Blend'' ($\rho=0.003$), ``Dynamic'' ($\rho=0.003$) and ``TaCT'' ($\rho_c=0.003$ and $\rho_p=0.003$).

As Table~\ref{tab:gtsrb_results} tells, considerable amount of our adaptive poison samples could still survive the latent separability based defenses on GTSRB, as on CIFAR-10. Furthermore, compared with the non-adaptive attacks, our adaptive attacks consistently perform better (smaller elimination rates).

\begin{table}[!t]
\centering
\caption{Results of our adaptive attacks on GTSRB.}
\vspace{-0.7em}
\resizebox{0.99\linewidth}{!}{ %< auto-adjusts font size to fill line
% \begin{tabular}{@{}lccc@{}}
\begin{tabular}{ccccccccccccccc}
\toprule
\textbf{Defenses$\rightarrow$} &
\multicolumn{2}{c}{No Defense} &
\multicolumn{2}{c}{Spectral Signature} &
\multicolumn{2}{c}{Activation Clustering} & 
\multicolumn{2}{c}{SCAn} & 
\multicolumn{2}{c}{SPECTRE} \cr
\cmidrule(lr){2-3} \cmidrule(lr){4-5} \cmidrule(lr){6-7} \cmidrule(lr){8-9} \cmidrule(lr){10-11} \cmidrule(lr){12-13}
\textbf{Attacks} $\downarrow$ & ASR & CA & Elimination & Sacrifice & Elimination & Sacrifice & Elimination & Sacrifice & Elimination & Sacrifice \cr
\midrule
Dynamic & 100.0 & 98.0 & 96.2 & 17.8 & 75.9 & 3.4 & 84.8 & 3.2 & 0.0 & 0.1 \cr
\midrule
TaCT & 100.0 & 97.7 & 97.5 & 17.8 & 0.0 & 3.0 & 0.0 & 0.0 & 0.0 & 0.1 \cr
\midrule
% Blend & 91.9 & 97.7 & 65.8 & 17.9 & 0.0 & 4.1 & 0.0 & 2.4 & 0.0 & 0.1 \cr
Blend & 86.4 & 97.6 & 70.9 & 17.8 & 84.8 & 6.9 & 0.0 & 3.6 & 0.0 & 0.1 \cr
\midrule
\textbf{Adap-Blend} & 82.2 & 97.8 & 43.0 & 17.9 & 0.0 & 3.5 & 0.0 & 2.3 & 0.0 & 0.1 \cr
\midrule
BadNet & 99.5 & 97.8 & 100.0 & 25.2 & 97.7 & 2.3 & 99.2 & 0.5 & 100.0 & 0.3 \cr
\midrule
\textbf{Adap-Patch} & 63.1 & 97.8 & 74.4 & 25.4 & 0.0 & 0.3 & 35.6 & 4.6 & 0.0 & 0.1 \cr
\bottomrule

\end{tabular}
} % \resizebox

\label{tab:gtsrb_results}
\end{table}

% \begin{table}[ht]
% \centering
% \resizebox{0.99\linewidth}{!}{ %< auto-adjusts font size to fill line
% % \begin{tabular}{@{}lccc@{}}
% \begin{tabular}{ccccccccccccccccccccccc}
% \toprule
% \textbf{Defenses$\rightarrow$} &
% \multicolumn{2}{c}{No Defense} &
% \multicolumn{4}{c}{Spectral Signature} &
% \multicolumn{4}{c}{Activation Clustering} & 
% \multicolumn{4}{c}{SCAn} & 
% \multicolumn{4}{c}{SPECTRE} \cr
% \cmidrule(lr){2-3} \cmidrule(lr){4-7} \cmidrule(lr){8-11} \cmidrule(lr){12-15} \cmidrule(lr){16-19} \cmidrule(lr){20-23}
% \textbf{Attacks} $\downarrow$ & ASR & CA & Eli & Sac & ASR & CA & Eli & Sac & ASR & CA & Eli & Sac & ASR & CA & Eli & Sac & ASR & CA \cr
% \midrule
% Adap-Blend & 82.2 & 97.8 & 43.0 & 17.9 & \red{63.7} & 96.9 & 0.0 & 3.5 & \red{90.7} & 96.2 & 0.0 & 2.3 & \red{93.9} & 97.7 & 0.0 & 0.1 & \red{87.1} & 96.2 \cr
% \midrule
% Adap-Patch & 63.1 & 97.8 & 74.4 & 25.4 & 19.4 & 96.7 & 0.0 & 0.3 & \red{67.3} & 96.7 & 35.6 & 4.6 & \red{47.4} & 97.5 & 0.0 & 0.1 & \red{72.2} & 96.5 \cr
% \bottomrule

% \end{tabular}
% } % \resizebox
% \caption{
% \textbf{Results of our adaptive attacks on GTSRB.} ``Eli'' for elimination rate, ``Sac'' for sacrifice rate, ``ASR'' for attack success rate and ``CA'' for clean accuracy.}
% \label{tab:gtsrb_results}
% \end{table}

\subsubsection{Imagenette}
\label{appendix:imagenette}

To illustrate the effectiveness adaptive strategy on high-resolution inputs (e.g., 224x224), we evaluate our adaptive poisoning backdoor attacks on a 10-classes Imagenet~\citep{russakovsky2015imagenet} subset. Specifically, we take the commonly used Imagenette~\citep{imagenette} subset.

To successfully backdoor poison such a high-resolution dataset, selected triggers should usually be stronger (more evident visually). Specifically, for ``Adap-Blend'', we find the original blending trigger (Hellokitty) not strong enough, and replaced it with a random noise blending trigger (training poison alpha 0.15, test alpha 0.2; $\rho_p=0.003, \rho_c=0.003$); same trigger is used for ``Blend'' (alpha 0.2 all the time; $\rho_p=0.003$). We also enhance ``Adap-Patch'' by training trigger opacities and their poison rates (Fig~\ref{fig:poison-demo-adaptive-k-trigger-0}-\ref{fig:poison-demo-adaptive-k-trigger-1}, opacity=1.0; $\rho_p=0.01$, $\rho_c=0.02$). For comparison, we use a stronger patch trigger (``Firefox'', Fig\ref{fig:poison-demo-adaptive-k-trigger-1}) with $\rho_p=0.01$ (the BadNet trigger is not strong enough for backdoor injection).

Table~\ref{tab:imagenette_results} shows that our adaptive attacks successfully circumvent the latent separability based defenses. On Imagenette, some defenses (Activation Clustering and SCAn with standard hyperparameters) are ineffective -- they cannot cleanse any poison samples even for Blend and Firefox. Even so, we can still notice the superiority of our adaptive strategy over naive poisoning attacks (significantly smaller elimination rates against Spectral Signature and SPECTRE).

\begin{table}[ht]
\centering
\caption{Results of our adaptive attacks on Imagenette.}
\vspace{-0.7em}
\resizebox{0.99\linewidth}{!}{ %< auto-adjusts font size to fill line
% \begin{tabular}{@{}lccc@{}}
\begin{tabular}{ccccccccccccccc}
\toprule
\textbf{Defenses$\rightarrow$} &
\multicolumn{2}{c}{No Defense} &
\multicolumn{2}{c}{Spectral Signature} &
\multicolumn{2}{c}{Activation Clustering} & 
\multicolumn{2}{c}{SCAn} & 
\multicolumn{2}{c}{SPECTRE} \cr
\cmidrule(lr){2-3} \cmidrule(lr){4-5} \cmidrule(lr){6-7} \cmidrule(lr){8-9} \cmidrule(lr){10-11} \cmidrule(lr){12-13}
\textbf{Attacks} $\downarrow$ & ASR & CA & Elimination & Sacrifice & Elimination & Sacrifice & Elimination & Sacrifice & Elimination & Sacrifice \cr
\midrule
Blend & 93.1 & 90.1 & 21.4 & 4.4 & 0.0 & 3.2 & 0.0 & 0.0 & 100.0 & 0.1 \cr
\midrule
Adap-Blend & 60.8 & 89.7 & 7.1 & 4.4 & 0.0 & 3.1 & 0.0 & 0.0 & 0.0 & 0.4 \cr
\midrule
% K-Patch & 99.9 & 88.7 & 57.4 & 14.6 & 0.0 & 2.4 & 0.0 & 0.0 & 0.0 & 1.5 \cr
Firefox & 98.3 & 89.8 & 70.2 & 14.4 & 0.0 & 2.4 & 0.0 & 0.0 & 100.0 & 9.9 \cr
\midrule
Adap-Patch & 85.7 & 89.5 & 14.9 & 15.0 & 0.0 & 2.5 & 0.0 & 2.7 & 0.0 & 1.5 \cr
\bottomrule

\end{tabular}
} % \resizebox
\label{tab:imagenette_results}
\end{table}

% \begin{table}[ht]
% \centering
% \resizebox{0.99\linewidth}{!}{ %< auto-adjusts font size to fill line
% % \begin{tabular}{@{}lccc@{}}
% \begin{tabular}{ccccccccccccccc}
% \toprule
% \textbf{Defenses$\rightarrow$} &
% \multicolumn{2}{c}{No Defense} &
% \multicolumn{2}{c}{Spectral Signature~\cite{tran2018spectral}} &
% \multicolumn{2}{c}{Activation Clustering~\cite{chen2018activationclustering}} & 
% \multicolumn{2}{c}{SCAn~\cite{tang2021demon}} & 
% \multicolumn{2}{c}{SPECTRE~\cite{hayase21a}} \cr
% \cmidrule(lr){2-3} \cmidrule(lr){4-5} \cmidrule(lr){6-7} \cmidrule(lr){8-9} \cmidrule(lr){10-11} \cmidrule(lr){12-13}
% \textbf{Attacks} $\downarrow$ & ASR & CA & Elimination & Sacrifice & Elimination & Sacrifice & Elimination & Sacrifice & Elimination & Sacrifice \cr
% \midrule
% Firefox & 98.3 & 89.8 & 70.2 & 14.4 & 0.0 & 2.4 & 0.0 & 0.0 & 100.0 & 9.9 \cr
% \midrule
% Adap-Firefox & 41.3 & 89.8 & 10.6 & 15.0 & 0.0 & 8.8 & 0.0 & 0.0 & 43.6 & 4.3 \cr
% \midrule
% K-Patch & 99.9 & 88.7 & 57.4 & 14.6 & 0.0 & 2.4 & 0.0 & 0.0 & 75.5 & 7.5 \cr
% \midrule
% Adap-Patch & 85.7 & 89.5 & 14.9 & 15.0 & 0.0 & 2.5 & 0.0 & 2.7 & 69.1 & 6.8 \cr
% \bottomrule

% \end{tabular}
% } % \resizebox
% \caption{
% \textbf{Results of our adaptive attacks on Imagenette.} SPECTRE is given the oracle knowledge of the target class for better comparison.}
% \label{tab:imagenette_results}
% \end{table}

\subsection{Adaptive Attacks on Different Architectures}
\label{appendix:other_architectures}

We also consider other model architectures, e.g., VGG-16~\citep{simonyan2014very} and MobileNet-V2~\citep{sandler2018mobilenetv2}. We evaluate these architectures on CIFAR-10 against our adaptive attacks. Notice that these architectures have much larger latent spaces -- MobileNet-V2 has a latent space of dimension 1280, and VGG-16 has a latent space of dimension 512 (for comparison, 64 for ResNet-20). Therefore, we have to adapt some defense configurations. Specifically, SCAn could not finish computing for MobileNet-V2 after 2h, so we manually reduce the latent representation's dimension to 128 by PCA before SCAn starts processing.

Overall, we show that our adaptive attacks also effectively circumvent latent separation based defenses. In Table~\ref{tab:cifar10_other_arch_results}, we demonstrate our adaptive attacks with these network architectures on CIFAR-10. As shown, none of these latent separability based defenses could completely eliminate our backdoor poison samples and most of them completely fail.

\begin{table}[!t]
\centering
\caption{Results of our adaptive attacks on other network architectures.}
\vspace{-0.7em}
\resizebox{0.8\linewidth}{!}{ %< auto-adjusts font size to fill line
% \begin{tabular}{@{}lccc@{}}
\begin{tabular}{clcccc}
\toprule
\multirow{2}*{\textbf{Defenses}$\downarrow$} & \textbf{Archs}$\rightarrow$ & 
\multicolumn{2}{c}{VGG-16} &
\multicolumn{2}{c}{MobileNet-V2}\cr
\cmidrule(lr){3-4} \cmidrule(lr){5-6}
& \textbf{Attacks}$\rightarrow$ & \multicolumn{1}{c}{Adap-Blend} &
\multicolumn{1}{c}{Adap-Patch} & \multicolumn{1}{c}{Adap-Blend} &
\multicolumn{1}{c}{Adap-Patch} \cr
\cmidrule(lr){3-3} \cmidrule{4-4} \cmidrule(lr){5-5} \cmidrule{6-6}

\multirow{2}*{\shortstack{Without Defense}}
& ASR & 74.0 & 63.6 & 72.2 & 96.4 \cr
& Clean Accuracy & 93.4 & 93.6 & 92.2 & 92.3 \cr
\midrule

\multirow{2}*{\shortstack{Spectral Signature}}
& Elimination Rate & 59.3 & 53.3 & 3.3 & 2.7 \cr
& Sacrifice Rate & 4.3 & 4.4 & 4.5 & 4.5 \cr
% & ASR & 17.6 & \red{33.1} & \red{65.3} & \red{91.6} \cr
% & Clean Accuracy & 92.2 & 92.6 & 91.6 & 91.7 \cr
\midrule

\multirow{2}*{\shortstack{Activation Clustering}}
& Elimination Rate & 0.0 & 0.0 & 0.0 & 0.0 \cr
& Sacrifice Rate & 0.0 & 0.0 & 0.0 & 0.0 \cr
% & ASR & \red{74.0} & \red{63.6} & \red{75.8} & \red{96.4} \cr
% & Clean Accuracy & 93.4 & 93.6 & 91.9 & 92.3 \cr
\midrule

\multirow{2}*{\shortstack{SCAn}}
& Elimination Rate & 0.0 & 0.0 & 0.0 & 0.0 \cr
& Sacrifice Rate & 0.0 & 0.0 & 0.0 & 4.7 \cr
% & ASR & \red{74.0} & \red{63.6} & 8.4 & \red{96.4} \cr
% & Clean Accuracy & 93.4 & 93.6 & 90.7 & 92.3 \cr
\midrule

\multirow{2}*{\shortstack{SPECTRE}}
& Elimination Rate & 0.0 & 0.0 & 0.0 & 0.0 \cr
& Sacrifice Rate & 0.5 & 0.5 & 0.5 & 0.5 \cr
% & ASR & \red{67.5} & \red{73.7} & 11.5 & 8.8 \cr
% & Clean Accuracy & 93.4 & 93.2 & 92.1 & 92.0 \cr
\midrule

\end{tabular}
} % \resizebox
\label{tab:cifar10_other_arch_results}
\end{table}

\section{Other State-of-the-art Backdoor Attacks}

\begin{figure}[!t]
\centering
\begin{subfigure}{0.3\textwidth}
    \includegraphics[width=\textwidth]{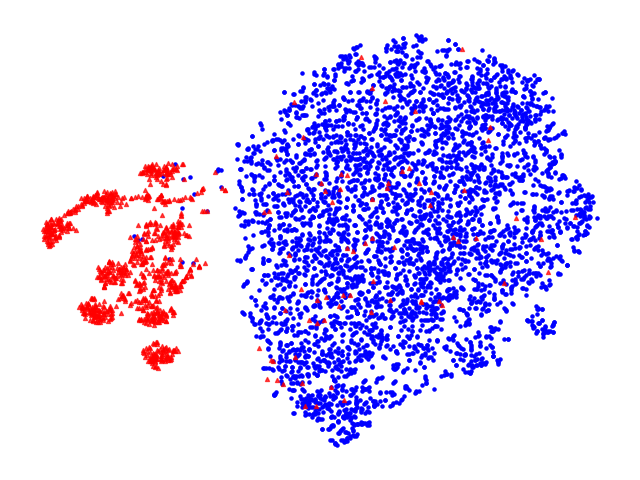}
    \caption{WaNet}
    \label{fig:wanet}
\end{subfigure}
\quad\quad
\begin{subfigure}{0.3\textwidth}
    \includegraphics[width=\textwidth]{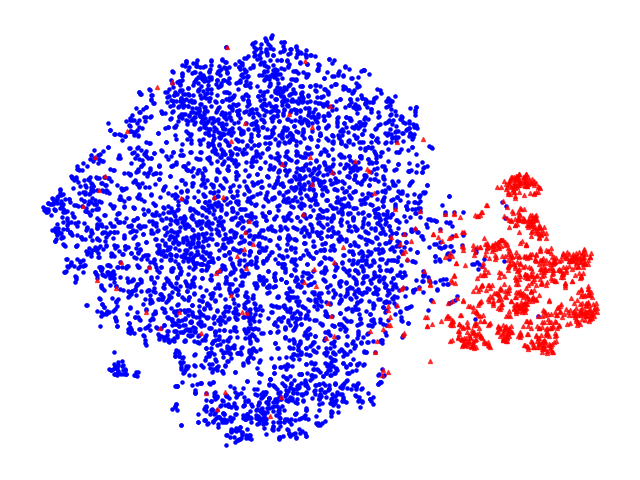}
    \caption{Refool}
    \label{fig:refool}
\end{subfigure}
\begin{subfigure}{0.3\textwidth}
    \includegraphics[width=\textwidth]{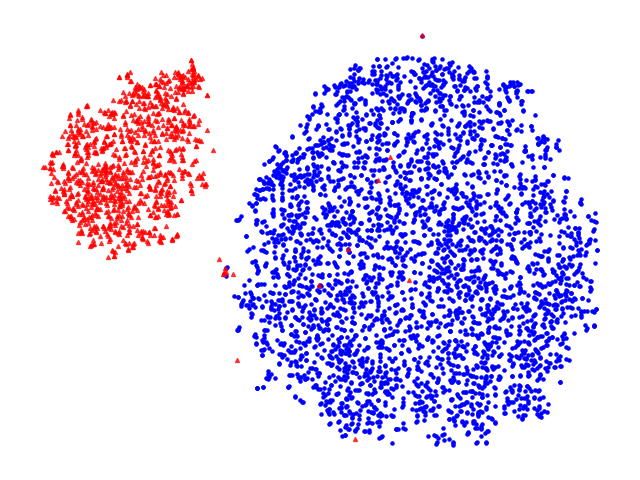}
    \caption{SIG}
    \label{fig:sig}
\end{subfigure}
\caption{t-SNE Visualization of latent separability of WaNet, Refool and SIG on CIFAR-10.}
\label{fig:wanet_and_refool}
\end{figure}

We also consider some other state-of-the-art backdoor attacks built on ``advanced triggers". We illustrate that they also suffer from latent separation. We take Refool~\citep{liu2020reflection}, WaNet~\citep{nguyen2021wanet} and SIG~\citep{barni2019new} as examples. Refool constructs realistic reflection triggers for backdoor injection, WaNet uses warp transformation for each image as the backdoor trigger, and SIG uses superimposed sinusoidal signals as triggers. We implement them in a poison-only manner, where they require much larger poison rates ($\rho=0.02$ for all three attacks) to inject backdoors with non-trivial ASR. %As shown in Tab~\ref{tab:wanet_and_refool}, latent separability based defenses can eliminate their poison samples by large fractions ($>80\%$). 
Their T-SNE visualizations in Fig~\ref{fig:wanet_and_refool} clearly reveals the latent separation.

\section{Other Defenses}

\begin{table}
\centering
\caption{The resistance of our adaptive attacks to other types of defenses on CIFAR-10. ``ASR'' for attack success rate, ``CA'' for clean accuracy, ``Eli'' for elimination rate, ``Sac'' for sacrifice rate, ``IP'' for isolation precision, ``STRIP (C) \& (F)'' for STRIP as a poison cleanser and an input filter.}
\vspace{-0.7em}
\resizebox{0.8\linewidth}{!}{ %< auto-adjusts font size to fill line
% \begin{tabular}{@{}lccc@{}}
\begin{tabular}{cccccccccccc}
\toprule
\textbf{Defenses$\rightarrow$} &
\multicolumn{2}{c}{FP} &
\multicolumn{2}{c}{STRIP (C)} & 
\multicolumn{2}{c}{STRIP (F)} & 
\multicolumn{2}{c}{NC} & 
\multicolumn{1}{c}{ABL} &
\multicolumn{2}{c}{NAD} \cr
\cmidrule(lr){2-3} \cmidrule(lr){4-5} \cmidrule(lr){6-7} \cmidrule(lr){8-9} \cmidrule(lr){10-10} \cmidrule(lr){11-12}
\textbf{Attacks} $\downarrow$ & ASR & CA & Eli & Sac & Eli & Sac & ASR & CA & IP & ASR & CA \cr
\midrule
Blend & 78.1 & 81.1 & 17.3 & 9.7 & 14.3 & 10.0 & 87.5 & 91.9 & 0.4 & 4.9 & 80.8  \cr
\midrule
\textbf{Adap-Blend} & 77.5 & 76.9 & 0.7 & 9.7 & 7.4 & 10.0 & 72.4 & 91.5 & 0.0 & 9.5 & 81.2 \cr
\midrule
BadNet & 88.9 & 80.7 & 100.0 & 10.1 & 100.0 & 10.0 & 1.7 & 90.3& 4.2 & 3.1 & 83.1 \cr
\midrule
\textbf{Adap-Patch} & 99.5 & 80.9 & 21.3 & 10.6 & 99.9 & 10.0 & 2.2 & 89.2 & 0.0 & 19.5 & 81.3 \cr
\bottomrule

\end{tabular}
} % \resizebox
\label{tab:cifar10_other_results}
\end{table}

% \begin{table}
% \centering
% \resizebox{0.8\linewidth}{!}{ %< auto-adjusts font size to fill line
% % \begin{tabular}{@{}lccc@{}}
% \begin{tabular}{clccccccc}
% \toprule
% \textbf{Defense} & ($\%$) & 
% \multicolumn{1}{c}{\textbf{Adaptive-Blend (Ours)}} &
% \multicolumn{1}{c}{\textbf{Adaptive-K (Ours)}}\cr
% \midrule
% % \cmidrule(lr){3-3} \cmidrule(lr){4-4}

% \multirow{2}*{\shortstack{Fine-Pruning~\cite{liu2018fine}}}
% & ASR & 48.1 & 72.3 \cr
% & Clean Accuracy & 80.8 & 79.3 \cr
% \midrule

% \multirow{4}*{\shortstack{STRIP (Cleanser)~\cite{gao2019strip}}}
% & Elimination Rate & 1.1 & 10.0 \cr
% & Sacrifice Rate & 10.3 & 10.5 \cr
% & ASR & 55.0 & 96.8 \cr
% & Clean Accuracy & 91.4 & 91.1 \cr
% \midrule

% \multirow{2}*{\shortstack{STRIP (Test-Time)~\cite{gao2019strip}}}
% & Elimination Rate & 1.2 & 100.0 \cr
% & Sacrifice Rate & 10.0 & 10.0  \cr
% \midrule

% \multirow{1}*{\shortstack{Neural Cleanse~\cite{wang2019neural}}}
% & Anomaly Index & 1.5 & 3.7\cr
% % & Anomaly Index & 1.2 & 1.7 & 2.9 & 2.9 & 1.5 & 3.7\cr
% % & ASR & / & 12.6 & 1.1 & 1.5 & \underline{\textbf{25.8}} & 2.5 \cr
% % & Clean Accuracy & 89.4 & 89.3 & 91.0 & 89.8 & 89.6 & 90.7 \cr
% \midrule

% \multirow{3}*{\shortstack{Anti-Backdoor\\Learning~\cite{li2021anti}}}
% & Isolation Precision & 0.0 & 0.4 \cr
% & ASR & 41.7 & 91.9 \cr
% & Clean Accuracy & 80.2 & 88.1 \cr
% \bottomrule

% \end{tabular}
% } % \resizebox
% \caption{
% \textbf{Other defenses against our adaptive attacks on CIFAR-10.}}
% \label{tab:cifar10_fp_results}
% \end{table}

Beyond the latent-space defenses, there are also other types of backdoor defenses (e.g., \citep{liu2018fine, gao2019strip, wang2019neural, li2021anti, li2021neural, borgnia2021strong, li2021backdoor, borgnia2021dp, shen2021backdoor}) that are not explicitly based on the latent separability assumption. These defenses could potentially ease the threat brought by our adaptive attacks. For a more comprehensive evaluation, we also examine our adaptive attacks against (some of) them. Specifically, we consider:
\begin{itemize}
    \item Fine-Pruning (FP) ~\citep{liu2018fine}: a \textit{model-pruning-based} backdoor defense that claims when a model is fed with clean inputs, its dormant neurons are more likely to be responsible for the backdoor task. FP eliminates a model's backdoor by pruning these dormant neurons until a certain clean accuracy drop.
    \item STRIP ~\citep{gao2019strip}: an input-filtering-based backdoor defense based on the observation that when a poison sample is superimposed by clean samples, the predicted class confidence drops heavily. We consider STRIP both as a cleanser (STRIP~(C) detects poison training samples) and a test-time filter (STRIP~(F) detects poison test samples).
    \item Neural Cleanse (NC) ~\citep{wang2019neural}: a \textit{reverse-engineering-based} backdoor defense that restores triggers by optimizing on the input domain. The authors claim a class with its reversed trigger having an abnormally small norm is more possibly a poisoned target class. Quantitatively, it calculates an anomaly index for each class w.r.t. the reversed triggers' mask norm, where classes with anomaly index >2 are judged as poisoned targets (outliers). Then, the smallest abnormal reversed trigger is patched on a small clean set to unlearn the model's backdoor.
    \item Anti-Backdoor Learning (ABL) ~\citep{li2021anti}: a \textit{poison-suppression-based} backdoor defense that utilizes local gradient ascent to isolate 1\% suspected training samples with the smallest losses. The authors claim these samples are more possible to be poison samples and may help unlearn the backdoor.
    \item Neural Attention Distillation (NAD) ~\citep{li2021neural}: a \textit{distillation-based} backdoor defense that unlearn backdoor in a teacher-student distillation fashion. NAD first finetunes the backdoor model with a holdout clean set to obtain a ``teacher'' model, then uses this ``teacher'' model to teach the original backdoor model (``student'') to further unlearn the backdoor.
\end{itemize}

% As shown, our adapitve attacks withstand FP.
% Our adaptive poisoning strategy breaks this observation, since the model must decide whether to classify a poison sample to the target class by also its clean semantic, thus usually produces a more \textbf{conservative} confidence.

\begin{figure}
\centering
\begin{subfigure}{0.4\textwidth}
    \includegraphics[width=0.99\textwidth]{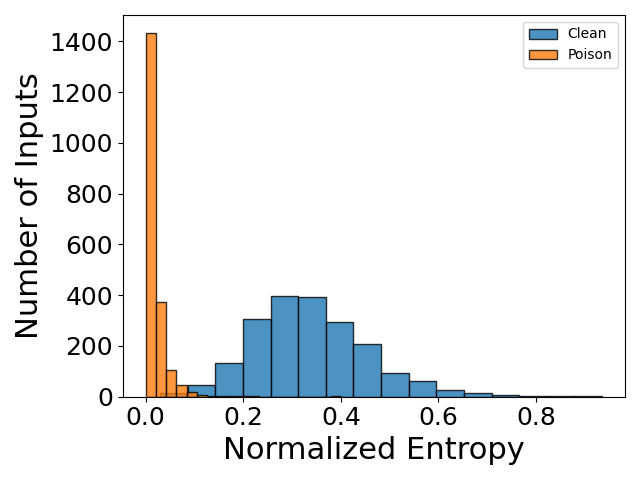}
    \label{fig:vis_strip_k_triggers_and_adaptive_k_0}
\end{subfigure}
\hspace{2em}
\begin{subfigure}{0.4\textwidth}
    \includegraphics[width=0.99\textwidth]{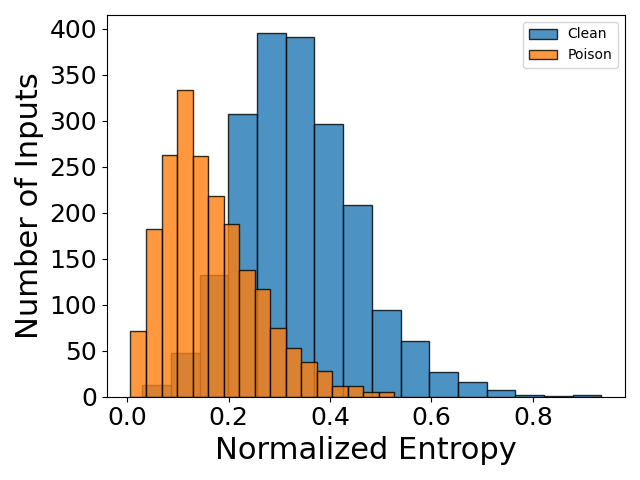}
    \label{fig:vis_strip_k_triggers_and_adaptive_k_1}
\end{subfigure}
\\
\begin{subfigure}{0.4\textwidth}
    \includegraphics[width=0.99\textwidth]{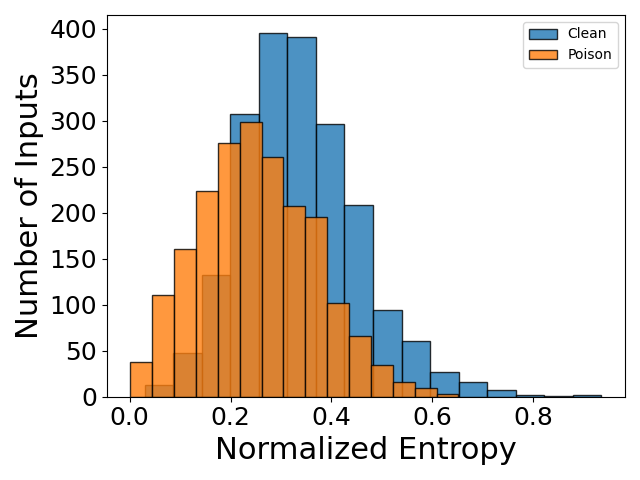}
    \label{fig:vis_strip_k_triggers_and_adaptive_k_2}
\end{subfigure}
\hspace{2em}
\begin{subfigure}{0.4\textwidth}
    \includegraphics[width=0.99\textwidth]{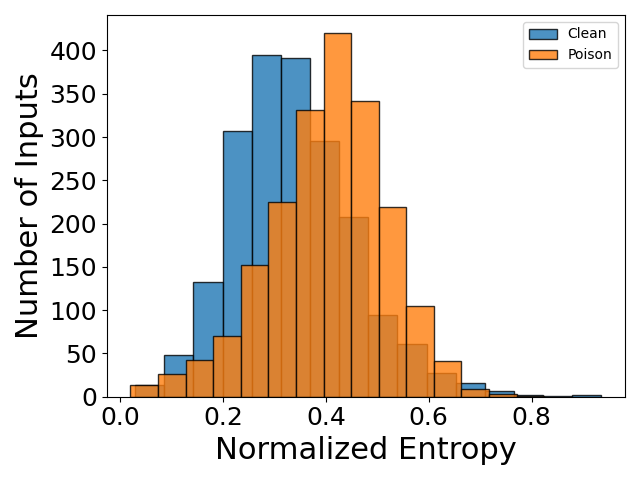}
    \label{fig:vis_strip_adap_patch_3}
\end{subfigure}
\vspace{-1em}
\caption{The STRIP normalized entropy histograms of Adap-Patch with different trigger selections. Samples with smaller entropy are more suspected to be poisoned. As shown, an attacker can achieve tradeoff between ASR and stealthiness by choosing different test-time triggers. Please refer to Table~\ref{tab:strip-tradeoff} for details of defense results.}
\label{fig:vis_strip_tradeoff}
\end{figure}

As shown in Table~\ref{tab:cifar10_other_results}: our adaptive methods do not decrease the resistance of their vanilla version to defenses other than latent-space ones. For STRIP~(C), our adaptive attacks even significantly outperform their vanilla version -- possibly due to the weaker correlation between the backdoor trigger and the target label in our poison samples. We also show in Table~\ref{tab:strip-tradeoff} and Figure~\ref{fig:vis_strip_tradeoff} that by selecting different test-time triggers, an Adap-Patch attacker can further evade the test-time STRIP (F) backdoor filter.

\begin{table}
\centering
\caption{
STRIP (as an input filter) against Adap-Patch attack with different trigger selections. The corresponding normalized histograms are shown in Fig~\ref{fig:vis_strip_tradeoff}.}
\vspace{-0.7em}
\resizebox{0.7\linewidth}{!}{ %< auto-adjusts font size to fill line
% \begin{tabular}{@{}lccc@{}}
\begin{tabular}{cccc}
\toprule
Triggers & ASR & Eli & Sac \cr
\midrule
\shortstack{Trigger~\ref{fig:poison-demo-adaptive-k-trigger-0} + Trigger~\ref{fig:poison-demo-adaptive-k-trigger-2}}
& 97.5 & 99.0 & 10.0 \cr
\hline
\shortstack{Trigger~\ref{fig:poison-demo-adaptive-k-trigger-1} + Trigger~\ref{fig:poison-demo-adaptive-k-poison-3}}
& 86.5 & 70.0 & 10.0 \cr
\hline
\shortstack{Trigger~\ref{fig:poison-demo-adaptive-k-trigger-0} (opacity=0.5) + Trigger~\ref{fig:poison-demo-adaptive-k-trigger-2} (opacity=0.7)}
& 59.7 & 35.0 & 10.0 \cr
\hline
\shortstack{Trigger~\ref{fig:poison-demo-adaptive-k-trigger-1}}
& 25.6 & 5.3 & 10.0 \cr
\bottomrule

\end{tabular}
} % \resizebox
\label{tab:strip-tradeoff}
\end{table}

\definecolor{myorange}{RGB}{255, 155, 0}
\definecolor{mypurple}{RGB}{117, 1, 117}

\begin{figure}
\centering
\includegraphics[width=.8\textwidth]{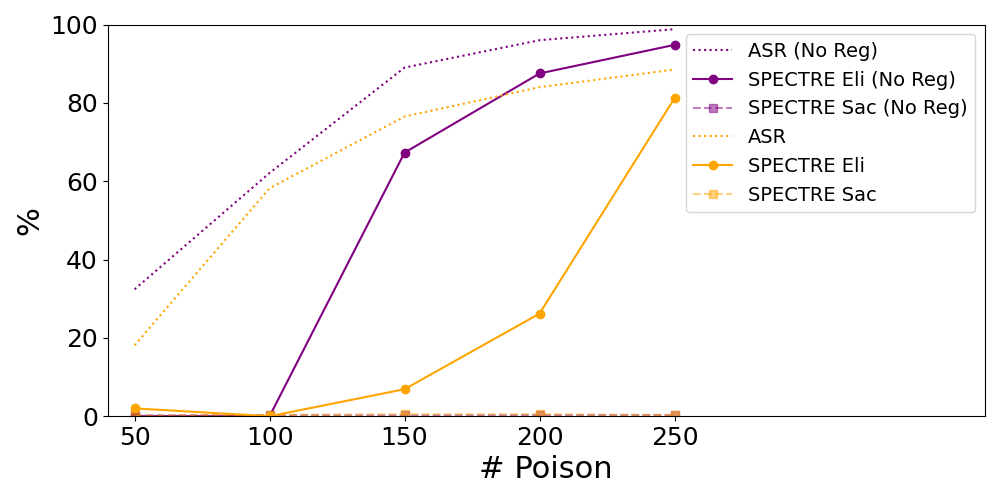}
\vspace{-0.8em}
\caption{Results of Adap-Blend with (\textcolor{myorange}{orange}) \& without (\textcolor{mypurple}{purple}) regularization samples. The dotted lines show the ASR. The solid lines correspond to Elimination  (``Eli'') and the dotted lines correspond to Sacrifice (``Sac''). As shown, Adap-Blend with regularization could remain stealthier in a larger range of poison rate selections (elimination $<30\%$ for 0-200 payload samples), compared to Adap-Blend without regularization that only uses our asymmetric triggers (elimination $>65\%$ for 150+ payload samples). \textbf{This further confirms the necessity of regularization.} }
\label{fig:ablation_no_reg}
\end{figure}

\section{Quantification of Latent (In)Separability}

% Although latent space visualizations already provide evaluations of separability that conform with human intuitions, we further design a numeric quantification metric for latent separability. Instead of commonly adopted unsupervised clustering metrics, we argue that fitting the training data's latent representations using a supervised SVM (with oracle knowledge of which of them are poisoned) could better indicate the latent separability.

% To quantify the fitting result of such a SVM, we could plot the Receiver Operating Characteristic (ROC) curve~\cite{hanley1982meaning} and calculate the Area Under the ROC Curve (AUC). AUC reflects how good the SVM could correctly classify the representations (into poison and clean); AUC $\in[0,1]$, where a larger AUC implies the better the SVM have separated the representations, and thus the less separability in latent space. So we quantify the Latent InSeparability (LIS) as:
% \begin{align}
%     LIS = (1 - AUC) * 1000 \ge 0
% \end{align}
% , where a larger LIS indicates that the backdoor poisoning attack is less separable in the latent space, vice versa. Actually, when LIS$ = 0$, the poison samples are totally separable in the latent space by an oracle SVM, which means the best defender who linearly analyzes the latent space could possibly remove all (and only remove) poison samples; and when LIS $ > 0$, even the best defender who linearly analyzes the latent space could not accurately remove all poison samples.

\textbf{Silhouette Score (Coefficient)} we show in Table ~\ref{tab:cifar10_main_results} formally measures such a property. For our case, Silhouette Score (Coefficient) measures how different are two labeled clusters ($i.e.$, knowing which samples are poisoned). The Silhouette Coefficient is calculated using the mean intra-cluster distance ($a$) and the mean nearest-cluster distance ($b$) for each sample. The Silhouette Coefficient for a sample is $(b - a) / \max(a, b)$. ($b$ is the distance between a sample and the nearest cluster that the sample is not a part of; $i.e.$, the other cluster in our case.) As shown in the last row of Table~\ref{tab:cifar10_main_results}, our Adap-attacks have significantly lower Silhouette Scores, which means the clean and poison clusters overlap more (and are thus more indistinguishable).

\section{The Comparison with LSBA}

We notice a concurrent work~\citep{peng2022label} proposing Label-Smoothed Backdoor Attack (LSBA) that shares certain similarity with ours adaptive backdoor attacks. We here list several fundamental differences between our work and LSBA: \textbf{1)} \textbf{Motivation}. Our method is motivated by our observations that the latent separability is so pervasive across both classical and advanced backdoor poisoning attacks. \cite{peng2022label} was inspired by the overfitting problem of existing backdoor attacks. \textbf{2)} \textbf{The Generation of Regularization/Poisoned Samples}. In general, both our method and \cite{peng2022label} preserve the ground-truth label (instead of re-assigning to the target label) of a fraction of modified samples. However, the fraction of our method is sample-independent whereas that of \cite{peng2022label} is sample-specific. This difference is because we have different mechanisms: we intend to alleviate the latent separability whereas they aimed to assign a soft-label to each poisoned sample. \textbf{3) The Improvement Module}. In general, both our method and \cite{peng2022label} have an additional module to further improve the main attack. However, we adopt asymmetric triggers to alleviate latent separability and encourage trigger diversity while preserving high attack effectiveness, whereas \cite{peng2022label} implanted multiple different backdoors to improve attack effectiveness. \textbf{4) Baseline Defenses}. We evaluate our attacks against latent separation based defenses, because our goal is to design adaptive backdoor poisoning attacks against this family of defenses. In contrast, \cite{peng2022label} only adopted STRIP~\citep{gao2019strip} and Neural Cleanse~\citep{wang2019neural} for evaluation. These two defenses are not based on latent separation.

Moreover, our method has unique contributions:
\begin{enumerate}
    \item  We formulate and reveal the latent separability assumption that is underlying many state-of-the-art defenses. We demonstrate that the latent separability assumption holds across a diverse set of backdoor poisoning attacks in the existing literature. Our work highlights the big blank in designing backdoor poisoning attacks that are stealthy in the latent representation space, and might consequently motivate future research in this direction. 
    \item In this work, we have done a proof-of-concept study to show that the latent separability assumption could fail. This conclusion suggests defense designers take caution when leveraging latent separation as an assumption in their defenses. We believe this can influence the design of backdoor defenses in the future. 
\end{enumerate}

\section{The Comparison with TaCT}

TaCT~\citep{tang2021demon} also uses a regularization technique. Specifically, TaCT injects regularization samples from several ``cover classes'' to ensure that only images from a ``source class'' are able to trigger the model's backdoor behavior when planted with the backdoor trigger. Whilst, our adaptive attack does not presume such ``source class'', and our regularization strategy is to randomly select regularization samples from all classes.

We here provide two extra ablation experiments to show that TaCT can hardly evade latent-space defenses, even when: \textbf{1)} injecting fewer poison samples (we explore the cases when there are only 50, 100, and 150 payload samples), and \textbf{2)} adopting our asymmetric trigger design (we use the same $k=4$ triggers from Adap-Patch following TaCT's regularization labelling strategy, denoted as \texttt{TaCT-k}). We show the defense results by SPECTRE in Tab~\ref{tab:ablation_TaCT}.

\begin{table}[!t]
\vspace{-2em}
\centering
\caption{Results of TaCT and TaCT-k with fewer payload samples. We use $\rho_\text{regularization}=\rho_\text{payload}$ for TaCT. for TaCT-k, we use $\rho_\text{regularization}=2 \cdot \rho_\text{payload}$. We only show defense results of SPECTRE since it consistently provides the highest poison elimination.}
\vspace{-0.7em}
\resizebox{0.6\linewidth}{!}{ %< auto-adjusts font size to fill line
\begin{tabular}{llccc}
\toprule
Attack$\downarrow$ & \# Payload Samples $\rightarrow$ & 50 & 100 & 150 \cr
\midrule
\multirow{4}*{TaCT}
& ASR (w/o defense) & 93.9 & 96.9 & 95.1 \cr
& Clean Acc (w/o defense) & 91.8 & 91.5 & 91.7 \cr
& SPECTRE Elimination Rate & 100.0 & 100.0 & 100.0 \cr
& SPECTRE Sacrifice Rate & 0.1 & 0.1 & 0.2 \cr

\midrule
\multirow{4}*{TaCT-k}
& ASR (w/o defense) & 95.4 & 100.0 & 100.0 \cr
& Clean Acc (w/o defense) & 92.0 & 91.9 & 91.8 \cr
& SPECTRE Elimination Rate & 90.0 & 95.0 & 99.3 \cr
& SPECTRE Sacrifice Rate & 0.1 & 0.1 & 0.2 \cr

\midrule
\multirow{4}*{\textbf{Adap-Blend}}
& ASR (w/o defense) & 18.1 & 58.2 & 76.5 \cr
& Clean Acc (w/o defense) & 91.8 & 91.4 & 91.6 \cr
& SPECTRE Elimination Rate & 2.0 & 0.0 & 6.9 \cr
& SPECTRE Sacrifice Rate & 0.1 & 0.3 & 0.5 \cr

\midrule
\multirow{4}*{\textbf{Adap-Patch}}
& ASR (w/o defense) & 75.1 & 83.0 & 97.5 \cr
& Clean Acc (w/o defense) & 91.6 & 91.8 & 91.9 \cr
& SPECTRE Elimination Rate & 6.0 & 0.0 & 0.0 \cr
& SPECTRE Sacrifice Rate & 0.1 & 0.3 & 0.5 \cr

\bottomrule
\end{tabular}
} % \resizebox
\label{tab:ablation_TaCT}
\end{table}

Clearly, even with fewer poison samples and the asymmetric k-triggers, TaCT (TacT-k) cannot evade latent-space detection ($\ge90\%$ payload samples are cleansed by SPECTRE). On the other hand, both our adaptive attacks can hardly be detected (only $<7\%$ payload samples are cleansed by SPECTRE), and thus significantly outperform TaCT $w.r.t.$ stealthiness against latent space defenses.

\section{The Comparison with M-Way Attack}

\cite{xie2019dba} proposes M-Way attack as an instance of distributed backdoor attack in federated learning scenarios, where $m$ independent pixels are used as $m$ distributed backdoor triggers, respectively. They proposes DBA targeting federated learning scenarios, and also studies a type of asymmetric triggers for data poisoning and testing time attacks respectively. Compared with our work, (1) M-Way attack still suffers from latent separation; (2) it doesn't consider regularization in data poisoning as well as its implication for suppressing latent separation characteristics.

To highlight the differences, we:
\begin{enumerate}
    \item Supplement an additional evaluation on the M-Way attack on CIFAR10. We follow the original settings of \cite{xie2019dba}, and use the same number of payload samples to that of our main evaluation in Table~\ref{tab:cifar10_main_results}.
    \item Further implement an adaptive version of the M-Way attack (Adap-M-Way) via incorporating our idea of regularization samples. Number of payload and regularization samples are also the same to that of Adaptive-Blend in Table~\ref{tab:cifar10_main_results}.
\end{enumerate}

As shown in Table~\ref{tab:ablation_m_way}, the original M-Way attack proposed by \cite{xie2019dba} can be defended (ASR$<5\%$) by 3 out of 4 latent separation-based defenses that we evaluate against. After being adapted to our regularization techniques, none of the four attacks can defend it.

\begin{table}[!t]
\centering
\caption{Additional results of comparing M-Way and Adap-M-Way attacks.}
\vspace{-0.7em}
\resizebox{0.55\linewidth}{!}{ %< auto-adjusts font size to fill line
% \begin{tabular}{@{}lccc@{}}
\begin{tabular}{clcc}
\toprule
\textbf{Defenses}$\downarrow$ & \textbf{Attacks}$\rightarrow$ & 
M-Way & \textbf{Adap-M-Way} \cr
\cmidrule(lr){3-3} \cmidrule{4-4}

\multirow{2}*{\shortstack{Without Defense}}
& ASR & 87.7 & 73.8 \cr
& Clean Accuracy & 91.3 & 91.0 \cr
\midrule

\multirow{4}*{\shortstack{Spectral Signature}}
& Elimination Rate & 33.3 & 19.0 \cr
& Sacrifice Rate & 4.4 & 4.5 \cr
& ASR & 4.7 & 66.6 \cr
& Clean Accuracy & 91.3 & 90.9 \cr
\midrule

\multirow{4}*{\shortstack{Activation Clustering}}
& Elimination Rate & 0.0 & 0.0 \cr
& Sacrifice Rate & 0.0 & 0.0 \cr
& ASR & 87.8 & 73.8 \cr
& Clean Accuracy & 91.3 & 91.0 \cr
\midrule

\multirow{4}*{\shortstack{SCAn}}
& Elimination Rate & 55.3 & 0.0 \cr
& Sacrifice Rate & 7.5 & 0.0 \cr
& ASR & 3.3 & 73.8 \cr
& Clean Accuracy & 91.0 & 91.0 \cr
\midrule

\multirow{4}*{\shortstack{SPECTRE}}
& Elimination Rate & 64.0 & 10.0 \cr
& Sacrifice Rate & 0.3 & 0.4 \cr
& ASR & 2.4 & 62.5 \cr
& Clean Accuracy & 91.4 & 91.5 \cr
\midrule

\end{tabular}
} % \resizebox
\label{tab:ablation_m_way}
\end{table}

\end{document}